\documentclass[sigconf]{acmart}
\AtBeginDocument{%
  }

\setcopyright{none}
\renewcommand\footnotetextcopyrightpermission[1]{}
\acmConference[]{}{}{}
\usepackage{graphicx}	
\usepackage{bbding}	
\usepackage{stfloats}
\usepackage{booktabs}
\usepackage{makecell}

\usepackage{graphics}
\usepackage{microtype}
\usepackage{epsfig}
\usepackage{caption}
\usepackage{float}
\usepackage{placeins}
\usepackage{color, colortbl}
\usepackage{enumitem}
\usepackage{tabularx}
\usepackage{xstring}
\usepackage{multirow}
\usepackage{xspace}
\usepackage{url}
\usepackage{subcaption}
\usepackage{xcolor}

\def\ie{{\em i.e.}}
\def\eg{{\em e.g.}}
\definecolor{myred}{RGB}{220, 38, 38}
\definecolor{mygreen}{RGB}{16, 185, 129}
\definecolor{myblue}{RGB}{37, 99, 235}

\definecolor{1st}{RGB}{220, 38, 38}   % vivid red
\definecolor{2nd}{RGB}{16, 185, 129}  % teal green
\definecolor{3rd}{RGB}{37, 99, 235}   % strong blue
\definecolor{top1p}{RGB}{183, 183, 235}
\definecolor{totalp}{RGB}{234, 184, 131}
\begin{document}

\title{To Blend In, First Decouple: Rethinking Camouflage Image Generation via Context-Decoupled Representations}
\author{Wenzhuang Wang}
\affiliation{%
  \institution{State Key Laboratory of Virtual Reality Technology and Systems, SCSE \& QRI, Beihang University}
  \city{Beijing}
  \country{China}
}
\email{wz\_wang@buaa.edu.cn}

\author{Yifan Zhao}
\authornote{Correspondence should be addressed to Yifan Zhao and Jia Li.}
\affiliation{%
  \institution{State Key Laboratory of Virtual Reality Technology and Systems, SCSE \& QRI, Beihang University}
  \city{Beijing}
  \country{China}
}
\email{zhaoyf@buaa.edu.cn}

\author{Mingcan Ma}
\affiliation{%
  \institution{Geely Automobile Research Institute (Ningbo) Co., Ltd, AI Center, Geely}
  \city{Ningbo}
  \country{China}
}
\email{Mingcan.Ma@geely.com}

\author{Yunlong Che}
\affiliation{%
  \institution{Geely Automobile Research Institute (Ningbo) Co., Ltd, AI Center, Geely}
  \city{Ningbo}
  \country{China}
}
\email{blizzak@163.com}

\author{Haoran Chen}
\affiliation{%
  \institution{State Key Laboratory of Virtual Reality Technology and Systems, SCSE \& QRI, Beihang University}
  \city{Beijing}
  \country{China}
}
\email{haoranchen@buaa.edu.cn}

\author{Ming Liu}
\affiliation{%
  \institution{Geely Automobile Research Institute (Ningbo) Co., Ltd, AI Center, Geely}
  \city{Ningbo}
  \country{China}
}
\email{liuming5@geely.com}

\author{Jia Li*}
\affiliation{%
  \institution{State Key Laboratory of Virtual Reality Technology and Systems, SCSE \& QRI, Beihang University}
  \city{Beijing}
  \country{China}
}
\email{jiali@buaa.edu.cn}

\renewcommand{\shortauthors}{}

\begin{abstract}
  Camouflage image generation (CIG) focuses on generating visually concealed objects that seamlessly blend into their backgrounds. Existing methods typically follow either background-guided paradigms that adapt object appearance via style transfer, or foreground-guided strategies that outpaint surrounding regions conditioned on object features. 
  However, they still suffer from appearance discrepancy and background artifacts. We attribute these limitations to cross-context representation leakage, where object and background cues are entangled in a coupled conditional space, resulting in ambiguous control and degraded camouflage fidelity. To tackle this, we propose a new context-decoupled generative paradigm, termed CamoDreamer, which aims to isolate contextual conditional guidance and explicitly decouple latent camouflage features into coordinated object and background control streams. First, a Contrast-aware Contextual Bridge is designed to model cross-context discrepancies and construct contrast-aware dual conditional guidance. Second, Context-Decoupled Assimilation Streams are employed to separate generative interactions conditioned on the dual guidance, while facilitating background rendering with target-aware cues in the latent space. Finally, a Frequency-Adaptive Contextual Blend module integrates complementary high-frequency textures and low-frequency structures from decoupled features to improve holistic coherence. Extensive experiments demonstrate that CamoDreamer consistently outperforms existing methods with a substantial margin, while maintaining a relatively lightweight design.
\footnote{Project page: https://github.com/fayewong666999/CamoDreamer}
\begin{figure}[!t]
\centering
\includegraphics[width=0.49\textwidth]{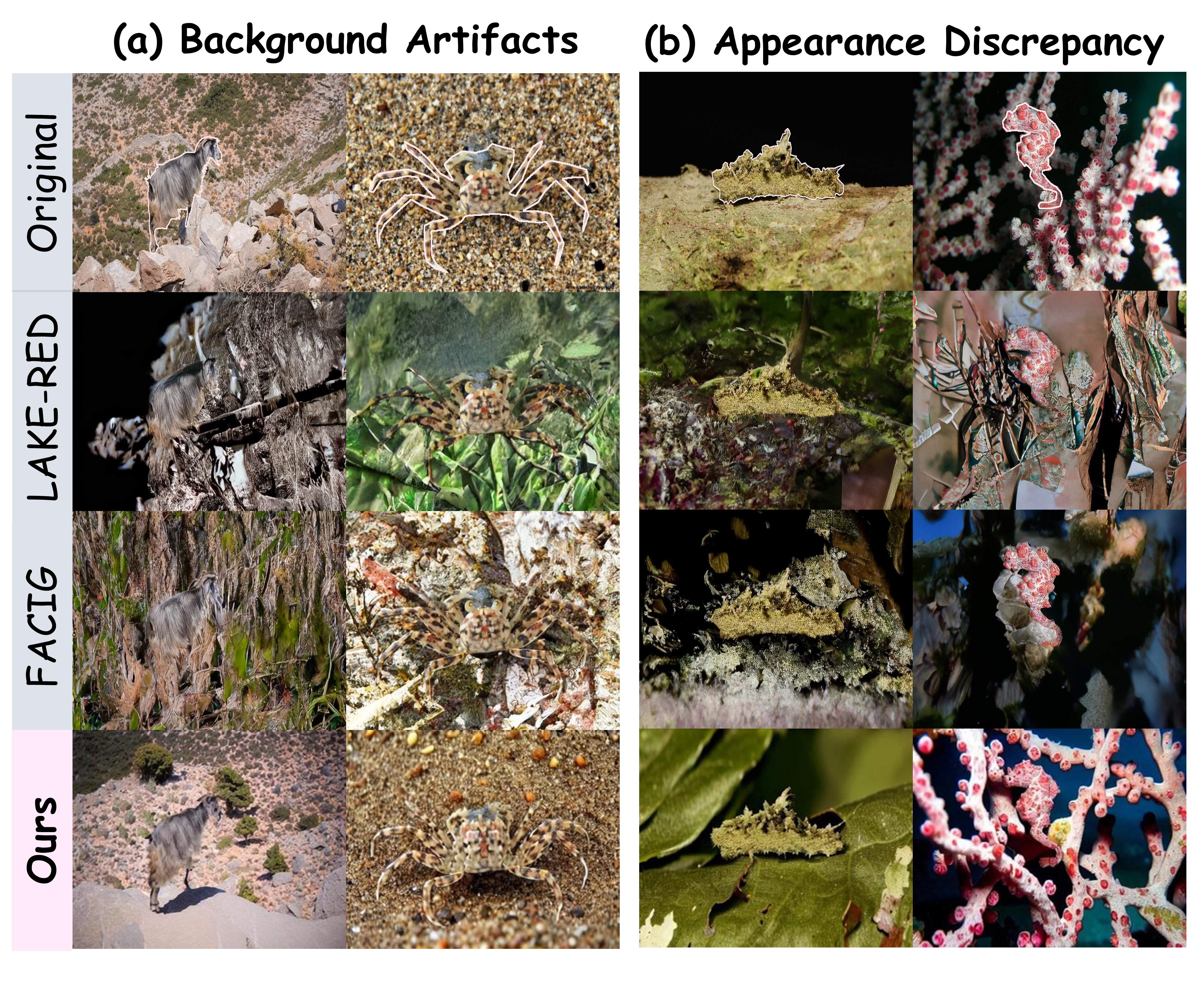}
  \caption{Motivation and qualitative comparison between our context-decoupled CamoDreamer and foreground-guided methods (LAKE-RED~\cite{zhao2024lake} and FACIG~\cite{chen2025foreground}). (a) Background artifacts: existing SOTA methods often produce unrealistic surroundings with severe texture artifacts. (b) Appearance discrepancy: noticeable visual inconsistency between foreground and background (\ie, color mismatch).}
  \label{fig:teaser}
\end{figure}
\end{abstract}


%\keywords{Cross-Context Representation Leakage; Camouflage Image Generation; Context-Decoupled Generative Paradigm}

\maketitle

\section{Introduction}
\textbf{$\bullet$ Background.} Camouflage is an evolutionary adaptation shaped by ``survival of the fittest''~\cite{stevens2009animal}, enabling organisms to remain highly consistent with their surroundings and avoid detection by predators. Given the inherent difficulty of identifying concealed objects and their distinct characteristics from generic images, camouflage visual recognition (\ie, camouflage object detection (COD)~\cite{xie2023frequency, he2024text, zhang2025frequency}) has attracted increasing attention for real-world applications, including medical diagnosis~\cite{xuan2025rethink, li2026m3net} and autonomous driving~\cite{li2024learning, huang2024alignsam, yang2025trajdiff}.

Unfortunately, progress in camouflage understanding is largely constrained by the scarcity of large-scale annotated datasets. For instance, the CAMO~\cite{le2019anabranch} dataset contains only about \textbf{2\%} of the images in the generic MS-COCO~\cite{lin2014microsoft}. This limitation mainly arises from the extremely intricate camouflage patterns, which make manual annotation labor-intensive and time-consuming, as well as the intrinsic rarity of camouflage in real-world scenes.

\noindent\textbf{$\bullet$ Limitations of Existing Methods.} With the rapid advancement of generative architectures (\ie, GANs~\cite{goodfellow2014generative, karras2019style} and diffusion models~\cite{ho2020denoising, rombach2022high}), recent efforts have leveraged their capabilities for camouflage image generation (CIG) to alleviate data scarcity. Existing CIG methods can be broadly categorized into two paradigms based on the guidance strategies: \textit{background-guided} and \textit{foreground-guided}. The former~\cite{chu2010camouflage, zhang2020deep, gao2025ptdiffusion}, exemplified by style transfer-based LCGNet~\cite{li2022location}, enables camouflage by modifying object color and texture to match the background. Nevertheless, the paradigm requires manually specified backgrounds and fails to preserve object appearance, leading to implausible contextual compositions. 

In contrast, the latter~\cite{lugmayr2022repaint, das2025camouflage, chen2025foreground} focuses on outpainting background regions conditioned on object representations while maintaining their semantic consistency. For example, LAKE-RED~\cite{zhao2024lake} exploits foreground features as queries to retrieve well-aligned backgrounds from a pre-trained codebook~\cite{van2017neural}, and combines them to facilitate background-matching camouflage generation. Despite enabling superficial concealment, foreground-guided methods still suffer from two major drawbacks: (1) \textit{background artifacts}, manifested as spurious texture details that compromise environmental realism, as illustrated by the distorted background of the crab in Fig.~\ref{fig:teaser}(a); and (2) \textit{appearance discrepancy}, \ie, color inconsistencies between generated backgrounds and target objects in Fig.~\ref{fig:teaser}(b).

\noindent\textbf{$\bullet$ Motivation and Our Effort.} Revisiting the camouflage generation process between foreground objects and their environments, we attribute these failures to \textit{\textbf{cross-context representation leakage}}, arising from entangled conditional guidance and coupled latent interactions. On the one hand, background knowledge is indiscriminately injected into foreground representations, causing foreground-dominant signals to overwhelm background rendering, as reflected by the severe artifacts in Fig.~\ref{fig:teaser}(a). On the other hand, the lack of structured interaction in the latent space leads to unreliable content generation, where incorrect background cues may interfere with the foreground, yielding visual discrepancies.

Motivated by this, our paper proposes a new \textit{context-decoupled} paradigm, termed \textbf{CamoDreamer}, which aims to factorize contextual guidance and explicitly decouple latent features into collaborative object and background streams to improve contextual compatibility. First, we design a Contrast-aware Contextual Bridge to learn object-to-background associations based on visual differences, and formulate contrast-aware dual conditional guidance for generation. Second, we introduce Context-Decoupled Assimilation Streams to separate the generative interaction between dual guidance and latent camouflage features to address control entanglement, where background rendering is enhanced with target-aware cues via an assimilation branch. Finally, to reinforce fine-grained textures and structural content, we develop a frequency-adaptive contextual blend module to  aggregate contextual high- and low-frequency camouflage features to ensure overall coherence.

\noindent\textbf{$\bullet$ Contributions} are threefold: (i) We present a new perspective on the failure of camouflage image generation, and identify \textit{cross-context representation leakage} as the fundamental cause. In light of this, we propose a novel \textbf{context-decoupled paradigm}, termed \textbf{CamoDreamer}, which factorizes contextual guidance into coordinated object and background control streams.
(ii) We devise a Contrast-aware Contextual Bridge to capture cross-context discrepancies and construct contrast-aware dual conditional guidance, alongside Context-Decoupled Assimilation Streams coordinating their interactions. In addition, a Frequency-Adaptive Contextual Blend module is developed to integrate complementary frequency cues for enhanced camouflage coherence.
(iii) Extensive experiments demonstrate that CamoDreamer consistently outperforms existing methods in fidelity and downstream perception.

\section{Related Work}
\textbf{Synthetic Data Generation.} Data generation has emerged as an effective strategy for augmenting training data in vision tasks~\cite{alimisis2025advances}, particularly in data-scarce scenarios. With the advent of advanced generative paradigms, such as GANs~\cite{goodfellow2014generative,zhang2017stackgan, zhu2019dm, tao2022df} and text-to-image (T2I) diffusion models~\cite{nichol2021glide, ramesh2022hierarchical, rombach2022high, saharia2022photorealistic, podell2023sdxl}, recent techniques harness their potential to synthesize large-scale datasets for downstream perception. DatasetGAN~\cite{zhang2021datasetgan} utilizes StyleGAN~\cite{karras2019style} to generate realistic images with pixel-wise labels at low cost, while BigDatasetGAN~\cite{li2022bigdatasetgan} further scales category diversity to ImageNet~\cite{deng2009imagenet} with limited supervision.  Most recently, diffusion-based methods, such as DiffuMask~\cite{wu2023diffumask} and DatasetDiffusion~\cite{nguyen2023dataset}, exploit pre-trained text-image alignment to produce realistic image-mask pairs without extra training. Moreover, DatasetDM~\cite{wu2023datasetdm} further improve annotation diversity by constructing a perceptual decoder in latent space. Despite these advances, existing methods focus on natural image synthesis and suffer domain gaps in intricate scenes.

\textbf{Controllable Image Generation.} Since T2I models often fail to capture fine-grained concepts (\ie, instance locations), controllable generation introduces spatial guidance, \ie, masks~\cite{lv2024place, li2025seg2any} and bounding boxes~\cite{yang2023reco, li2023gligen, chen2023geodiffusion, gu2025roictrl} to improve precise controllability. ControlNet~\cite{zhang2023adding} copies a trainable encoder from the denoising U-Net~\cite{ronneberger2015u} to incorporate diverse control signals. Building upon this, ControlNeXt~\cite{peng2024controlnext} adopts a lightweight design with cross-normalization, replacing zero-convolution operations for efficient conditioning. Instance-level control efforts~\cite{zhou2024migc, zhang2024cc, wang2025ficgen, bao2026envisioning} decompose spatial layouts into local regions to modulate object attributes, while ObjCtrl~\cite{zhang2025objctrl} leverages depth maps to model interactions between object semantics and latent features. Furthermore, training-free methods like BoxDiff~\cite{xie2023boxdiff} and LoCo~\cite{zhao2025loco}, manipulate latent attention maps with localized constraints to enable spatial control.

\textbf{Camouflage Image Generation.} Unlike generic image synthesis, CIG emphasizes contextual compatibility, requiring generated objects to blend into the surroundings. Early studies~\cite{chu2010camouflage} rely on hand-crafted features to mimic background patterns. Existing methods can be broadly categorized into two paradigms: \textit{background-guided} methods integrate objects into backgrounds via style transfer and structural alignment. For instance, DCI~\cite{zhang2020deep} designs an attention-aware camouflage loss to suppress object saliency, while LCGNet~\cite{li2022location} improves structural consistency via position-aligned fusion. However, they often overfit objects to local surroundings, delivering implausible content. In contrast, \textit{foreground-guided} methods synthesize backgrounds conditioned on given objects. LAKE-RED~\cite{zhao2024lake} pioneers a retrieval-based background outpainting strategy using a VQ-VAE codebook~\cite{van2017neural}, while later works~\cite{chen2025foreground, chen2025realcamo, das2025camouflage} improve feature integration and use text-visual guidance for perceptual consistency. CT-CIG~\cite{qian2026text} further exploits large vision-language models~\cite{wang2024qwen2} to obtain camouflage-aware descriptions. Nevertheless, these methods struggle with generative failures (\eg, background artifacts) due to entangled contextual control. To address this, we propose a context-decoupled paradigm that explicitly isolates object and background representations into independent rendering processes before blending.

\section{Methodology}
\begin{figure*}[!t]
\centering
\includegraphics[width=0.99\textwidth]{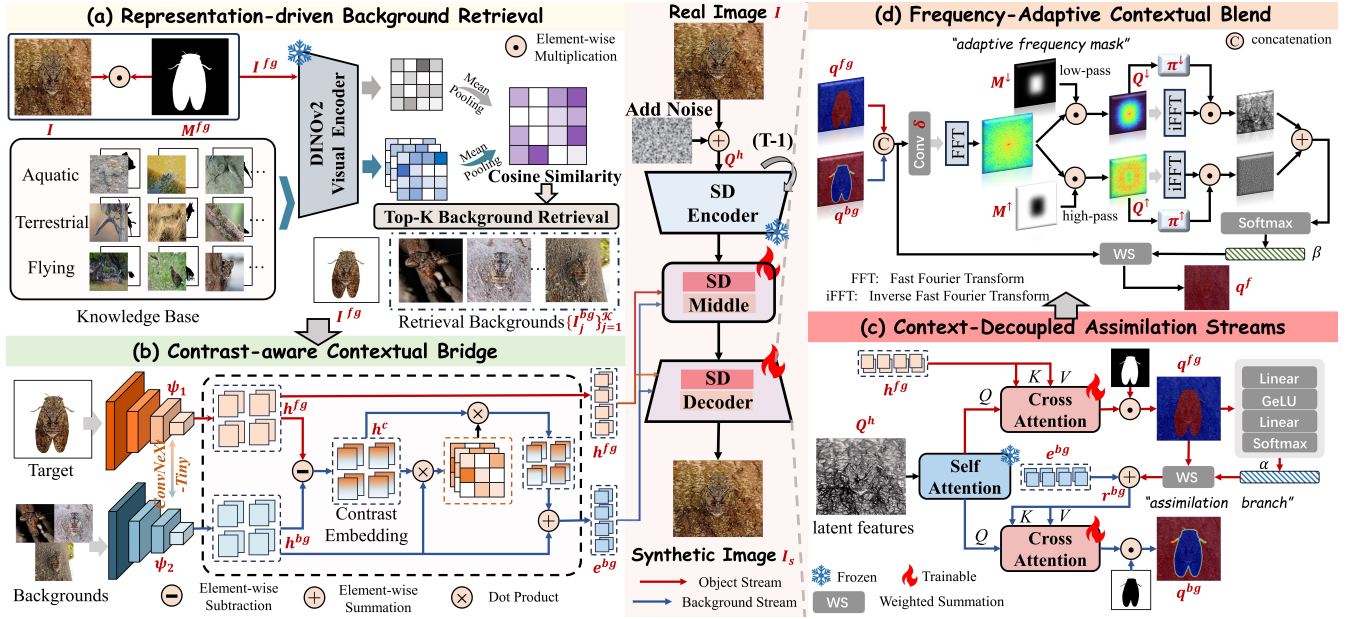}
\caption{\textbf{Overview of the proposed CamoDreamer framework}, which performs CIG via three major steps: (1) \textbf{a contrast-aware contextual bridge} that establishes the object-background associations for constructing dual conditional guidance; (2) \textbf{Context-Decoupled Assimilation Streams}, which decouple camouflage features into object and background streams, and then compensates the background with target-aware cues, and (3) \textbf{Frequency-Adaptive Contextual Blend} that dynamically integrates the decoupled features to ensure structural coherence.}
\label{corefigure}
\end{figure*}
\subsection{Preliminary}
\textbf{Stable Diffusion.} Following prior work~\cite{zhao2024lake, das2025camouflage}, our CamoDreamer is built upon the pre-trained Stable Diffusion (SD)~\cite{rombach2022high}. Compared with DDPM~\cite{ho2020denoising}, SD performs the Markov chain in a compact latent space, enabling a better trade-off between computation efficiency and image quality. Specifically, a camouflage image $\mathcal{I}$ is first encoded into a latent representation as $z_{0}$ via the VQ-VAE~\cite{van2017neural}, which is then perturbed by the forward diffusion into the noisy latents:
\begin{equation}
\label{eq1}
z_{t}=\sqrt{\overline{\alpha}_{t}}z_{0}+\sqrt{1-\overline{\alpha}_{t}}\epsilon, \quad \epsilon \sim \mathcal{N}(0,1),
\end{equation}
where $\overline{\alpha}_{t}=\prod_{i=1}^{t}\alpha_{i}$ and $\alpha_{t}=1-\beta_{t}$ denotes a predefined noise schedule, and $t\in \{1,T\}$ is the diffusion timestep. As $\overline{\alpha}_{t}$ decreases with $t$, increasing amounts of Gaussian noise $\epsilon$ are injected into $z_{0}$. Afterwards, a conditional denoising U-Net $\epsilon_{\theta}(\cdot)$~\cite{ronneberger2015u} is optimized to predict the injected noise $\epsilon$, with the objective function:
\begin{equation}
\label{eq2}\mathop{min}\limits_{\theta}\mathcal{L}_{LDM}=\mathbb{E}_{z_{0},\epsilon\sim \mathcal{N}(0,1),t,\mathcal{\tau}}[||\epsilon-\epsilon_{\theta}(z_{t},t,\mathcal{Y})||_{2}^{2}],
\end{equation}
where $\mathcal{Y}$ is the conditional guidance (\ie, text prompts).
\subsection{Framework Overview of CamoDreamer} 
To seamlessly blend generated objects into their backgrounds for camouflage compatibility, CamoDreamer follows a \textit{``first decouple, then blend''} paradigm that disentangles latent camouflage features into collaborative object and background control streams, and dynamically integrating them to mitigate cross-context representation leakage. 
As illustrated in Fig.~\ref{corefigure}, given a camouflage image $\mathcal{I}\in \mathbb{R}^{H\times W\times 3}$ and its binary mask $M^{fg}\in \mathbb{R}^{H\times W\times 1}$, we first retrieve a set of background images $\{\mathcal{I}_{j}^{bg}\}_{j=1}^{\mathcal{K}}$ by exploiting the representation capability of DINOv2~\cite{oquab2023dinov2}. 
Instead of directly compositing foreground and background, we introduce a Contrast-aware Contextual Bridge (CCB, $\mathcal{F}_{\text{CCB}}(\cdot)$) to model cross-context discrepancies and construct contrast-aware dual conditional guidance.
Building on this, the Context-Decoupled Assimilation Streams (CDAS, $\mathcal{F}_{\text{CDAS}}(\cdot)$) explicitly separate latent camouflage features $Q^{h}$ into foreground and background streams, while refining background synthesis via target-aware assimilation cues for visual consistency. 
Next, a Frequency-Adaptive Contextual Blend module (FACB, $\mathcal{F}_{\text{FACB}}(\cdot)$) integrates the decoupled $q^{fg}$ and $q^{bg}$ into a unified representation by reconciling high-frequency textures
and low-frequency structures. 
Finally, the VQ-VAE decoder $\mathcal{D}(\cdot)$ yields the synthetic image $\mathcal{I}_{s}$ from this unified representation.
The overall process can be formulated as:
\vspace{-2pt}
\begin{align}
\label{overall}
\text{Decouple:}\{q^{fg}, q^{bg}\} &= 
\mathcal{F}_{\text{CDAS}}(Q^{h}, \mathcal{F}_{\text{CCB}}(\mathcal{I}\odot M^{fg}, \{\mathcal{I}_{j}^{bg}\}_{j=1}^{\mathcal{K}})), \\
\text{Blend:}\,\mathcal{I}_{s} &= \mathcal{D}(\mathcal{F}_{\text{FACB}}(q^{fg},q^{bg})).
\end{align}
\subsection{Learning Contextual Associations}
As discussed above, CIG requires background rendering to be highly consistent with targets. To this end, existing methods such as LAKE-RED~\cite{zhao2024lake} and FACIG~\cite{chen2025foreground} utilize object features as queries to retrieve relevant background embeddings from a pre-trained VQ-VAE codebook. However, a notable domain gap exists between these generic embeddings and real-world camouflage scenarios, yielding unreliable object-to-background relation learning.

Hence, we leverage real camouflage images~\cite{chen2025realcamo} to derive the knowledge base, and introduce the representation-driven background retrieval to obtain domain-related background images as appearance anchors. Unlike prior studies that directly combine object and background features~\cite{zhao2024lake, chen2025realcamo}, we advocate the perception of representational differences between them as pivotal for understanding camouflage contexts, and design a contrast-aware context bridge to establish more informative correspondences.

\textbf{Representation-driven Background Retrieval.} Given a camouflage image $\mathcal{I}\in \mathbb{R}^{H\times W\times 3}$ and its object mask $M^{fg}\in\{0,1\}^{H\times W}$, we employ a self-supervised visual foundation model, \ie, DINOv2~\cite{oquab2023dinov2} ($\Phi(\cdot)$), to encode both the object image $\mathcal{I}^{fg}$ and candidate images in the knowledge base into a shared feature space. 
The Top-$\mathcal{K}$ most relevant camouflage images are then retrieved as background anchors based on cosine similarity:
\begin{equation}
\label{sim}
s_{i} = \frac{\Phi(\mathcal{I}\odot M^{fg})^{\top} \, \Phi(\mathcal{I}_{i}^{bg}\odot M_{i}^{bg})}
{\|\Phi(\mathcal{I}\odot M^{fg})\|_{2} \cdot \|\Phi(\mathcal{I}_{i}^{bg}\odot M_{i}^{bg})\|_{2}}, \quad i \in \{1,.., N\},
\end{equation}
where $\mathcal{I}^{fg}=\mathcal{I}\odot M^{fg}$, with $\odot$ denoting element-wise multiplication. 
$\mathcal{I}_{j}^{bg}$ and $M_{j}^{bg}$ denote the $j$-th candidate image and its corresponding binary background mask from a knowledge base of size $N$.
To this end, we formulate the retrieval set as $F = (\mathcal{I}^{fg}, \{\mathcal{I}_{j}^{bg}\}_{j=1}^{\mathcal{K}})$, which serves as contextual guidance for camouflage generation.

\textbf{Contrast-aware Contextual Bridge.} Instead of directly injecting background information around the target region, we emphasize their  contrasts to expose cross-context visual inconsistency. In this regard, we  employ trainable Siamese ConvNets ($\psi_{1,2}(\cdot)$) to extract camouflage features from the retrieval set. As shown in Fig.~\ref{corefigure}(a), for each object-background pair, a contrast embedding $h_{j}^{c}$ is computed to capture the distinctive characteristics of background anchors relative to the foreground. Subsequently, we design a gated cross-attention layer to update the background features conditioned on $h_{j}^{c}$, thus promoting the dominance of the background that is more similar to the target. The overall process is formulated as follows:
\begin{equation}
h_{j}^{c} = h_{j}^{bg} - h^{fg} \in \mathbb{R}^{L\times d_{c}}, \quad j\in \{1,\dots,\mathcal{K}\},
\end{equation}
\begin{equation}
e_{j}^{bg} = h_{j}^{bg} + \tanh(\gamma)\cdot
\mathrm{softmax}\!\left(
\frac{(h_{j}^{bg}W^{Q})(h_{j}^{c}W^{K})^{\top}}{\sqrt{d_{c}}}
\right)
h_{j}^{c}W^{V},
\end{equation}
where $h_{j}^{bg}=\psi_{1}(\mathcal{I}_{j}^{bg})$ and 
$h^{fg}=\psi_{2}(\mathcal{I}^{fg})$, both in $\mathbb{R}^{L\times d_{c}}$, 
with $L$ denoting the sequence length and $d_{c}$ the channel dimension. 
The encoders $\psi_{1,2}(\cdot)$ are instantiated as lightweight ConvNeXt-Tiny~\cite{liu2022convnet} to capture camouflage-specific color and texture patterns. 
$W^{Q}$, $W^{K}$, and $W^{V}$ denote learnable projection matrices, and $\gamma$ is a learnable scalar initialized to 0. Afterwards, the Top-$\mathcal{K}$ updated features $\{e_{j}^{bg}\}_{j=1}^{\mathcal{K}}$ are averaged across the retrieved candidates, \ie, $e^{bg} = \frac{1}{\mathcal{K}}\sum_{j=1}^{\mathcal{K}}e_{j}^{bg}$, to provide conditional guidance for the surrounding.
\begin{figure}[!t]
\centering
\includegraphics[width=0.48\textwidth]{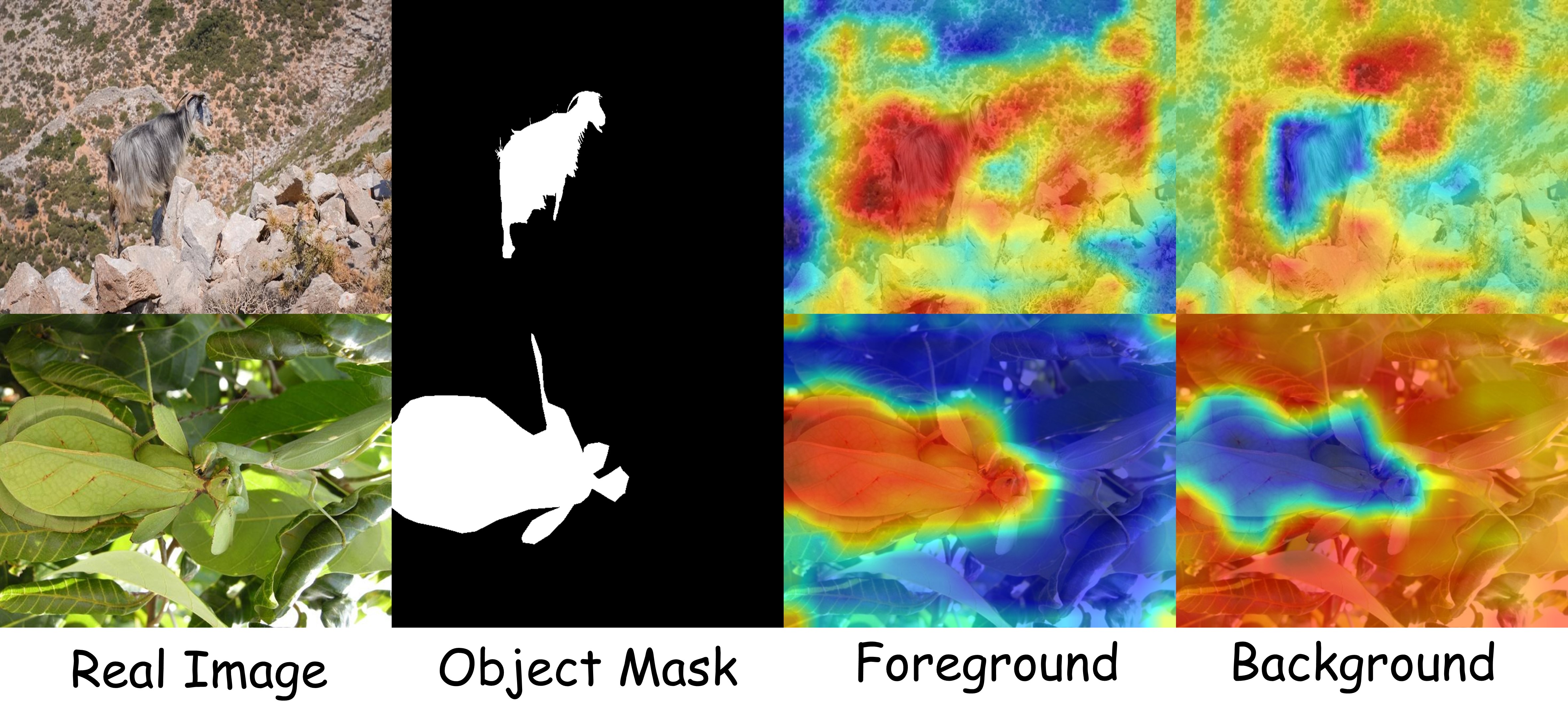}
\caption{Visualization of decoupled camouflage features ($q^{fg}$ and $q^{bg}$) at a spatial resolution of $16\times16$.}
\label{attnmap}
\end{figure}
\subsection{Context-Decoupled Assimilation Streams}
Regarding realistic camouflage generation, existing CIG methods typically depend on contextual conditioning via prompt engineering~\cite{das2025camouflage, qian2026text, chen2025realcamo} or feature integration~\cite{zhao2024lake, chen2025foreground}. Despite their straightforward designs, they often entangle foreground and background guidance within a coupled condition space, resulting in ambiguous control over contextual features in the latent space. Consequently, background features are easily contaminated by foreground signals, delivering noticeable texture artifacts, as illustrated in Fig.~\ref{fig:teaser} (a).

Therefore, to isolate contextual conditioning, we propose Context-Decoupled Assimilation Streams (CDAS) that explicitly disentangle latent camouflage features into independent object and background attention streams. As illustrated in Fig.~\ref{corefigure} (c), for the object stream, the feature $h^{fg}$ serves as conditional guidance to modulate the generation of camouflage object features $q^{fg}$ via cross-attention:
\begin{equation}
\label{foreattention}
q^{fg} = \mathrm{softmax}\left(
\frac{(Q^{h}W_{1}^{Q})(h^{fg}W_{1}^{K})^{\top}}{\sqrt{d}}
\right)
(h^{fg}W_{1}^{V}) \odot \overline{M}^{fg},
\end{equation}
where $Q^{h}\in \mathbb{R}^{L'\times d}$ denotes the latent camouflage features, $d$ is the latent dimension. $W_{1}^{Q}, W_{1}^{K}$, and $W_{1}^{V}\in \mathbb{R}^{d\times d}$ are learnable projection matrices. $\overline{M}^{fg}$ is the object mask (downsampled from $M^{fg}$), which enforces spatial constraints by restricting the attention to foreground regions, thus alleviating the representation leakage. For the background stream, beyond its inherent guidance, we introduce an assimilation branch that injects target cues for feature-level camouflage. Specifically,  we aggregate object features along the sequence dimension to obtain a compact target descriptor, which is then fused with $e^{bg}$ to facilitate target-aware background rendering:
\begin{equation}
r^{bg} = \sum_{l=1}^{L'} \alpha_l \, q_l^{fg} + e^{bg}, \quad 
\alpha = \mathrm{softmax}(\mathrm{MLP}(q^{fg})),
\end{equation}
\begin{equation}
q^{bg} = \mathrm{softmax}\left(
\frac{(Q^{h}W_{2}^{Q})(r^{bg}W_{2}^{K})^{\top}}{\sqrt{d}}
\right)
(r^{bg}W_{2}^{V}) \odot (1 - \overline{M}^{fg}),
\end{equation}
where $q^{bg} \in \mathbb{R}^{L' \times d}$ denotes the background features, and $\alpha \in \mathbb{R}^{L'}$ represents aggregation weights predicted by an MLP. 
The mask $(1 - \overline{M}^{fg})$ confines attention to background regions, effectively suppressing interference from foreground components. Fig.~\ref{attnmap} visualizes the intermediate camouflage feature maps at $16\times16$ resolution. With the proposed assimilation streams, foreground rendering concentrates on the target, while background rendering prioritizes semantically compatible regions (\ie, rocks and grass).
\begin{figure*}[!t]
\centering
\includegraphics[width=0.96\textwidth]{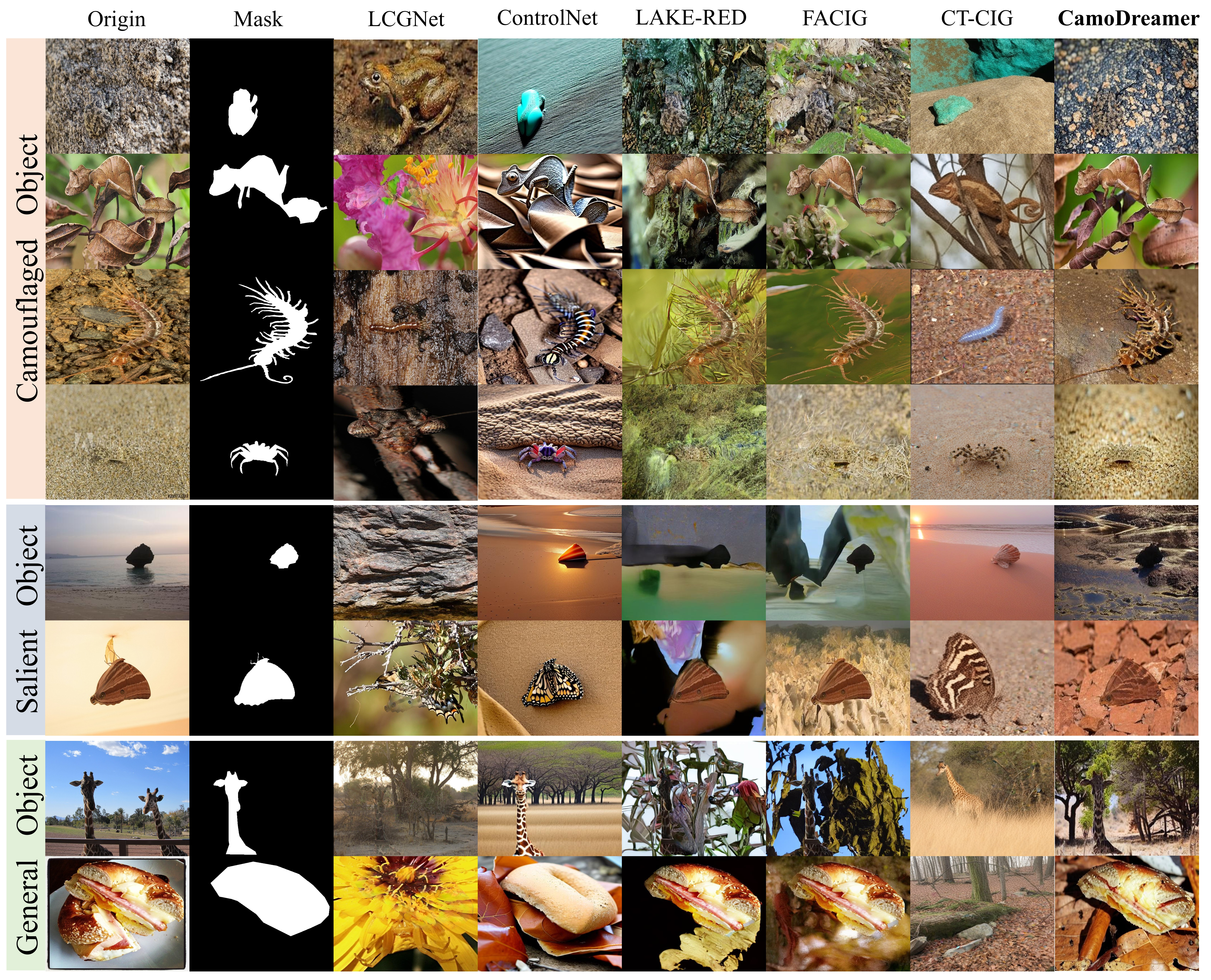}
\caption{Qualitative comparison of camouflage images generated by different CIG methods. The first two columns show real images and their masks. The background in the third column (LCGNet) is obtained via our representation-driven retrieval.}
\label{results}
\end{figure*}
\subsection{Frequency-Adaptive Contextual Blend}
After obtaining context-decoupled features, the final step is to integrate them into a unified representation for VQ-VAE decoding. Generally, foreground objects in camouflage scenes often exhibit rich high-frequency details, while backgrounds are dominated by low-frequency structures. Motivated by ~\cite{chen2024semantic}, we adopt a frequency-adaptive blending strategy to reconcile these complementary cues. 

As shown in Fig.~\ref{corefigure}(d), the decoupled features (\ie, $q^{fg}$ and $q^{bg}$) are first combined and transformed into the frequency domain via the Discrete Fourier Transform (DFT, $\mathcal{F}(\cdot)$). Adaptive frequency masks are then applied to isolate high- and low-frequency components:
\begin{equation}
    \textbf{Q}^{\uparrow} = \mathcal{F}(\delta([q^{fg}; q^{bg}])) \odot \textbf{M}^{\uparrow}, \quad
    \textbf{Q}^{\downarrow} = \mathcal{F}(\delta([q^{fg}; q^{bg}])) \odot \textbf{M}^{\downarrow},
\end{equation}
where $\delta(\cdot)$ denotes a convolutional projection. $\textbf{M}^{\uparrow}$ and $\textbf{M}^{\downarrow}$, where $\textbf{M}^{\downarrow}=1-\textbf{M}^{\uparrow}$, are adaptive high- and low-pass masks learned from the projected features $\delta([q^{fg}; q^{bg}])$. 
In particular, they are derived via global pooling followed by a linear layer and a sigmoid activation, which constrains the mask responses within $[0,1]$.

Subsequently, the frequency components are transformed back to the spatial domain and modulated by channel-wise attention:
\begin{equation}
    \{\beta_{1},\beta_{2}\} = \mathrm{softmax}\Big(
    \mathcal{F}^{-1}(\textbf{Q}^{\uparrow}) \odot \pi^{\uparrow}(\textbf{Q}^{\uparrow}) +
    \mathcal{F}^{-1}(\textbf{Q}^{\downarrow}) \odot \pi^{\downarrow}(\textbf{Q}^{\downarrow})
    \Big),
\end{equation}
where $\mathcal{F}^{-1}(\cdot)$ denotes the inverse DFT, and $\pi^{\uparrow}(\cdot)$ and $\pi^{\downarrow}(\cdot)$ denote channel attention modules for capturing frequency-aware camouflage cues. Finally, the blended camouflage representation is obtained via weighted summation:
\begin{equation}
\label{frequency1}
q^{f} = \beta_{1} \odot q^{fg} + \beta_{2} \odot q^{bg},
\end{equation}
where $\beta_{1}$ and $\beta_{2}$ are frequency-adaptive weighting maps. $q^{f}$ is the unified representation decoded into the synthetic camouflage image.
\begin{table*}[htbp]
\begin{center}
\caption{\textbf{Quantitative comparison with SOTA methods on the LAKE-RED dataset.} The \textcolor{myred}{\textbf{1$^{st}$}}, \textcolor{mygreen}{\textbf{2$^{nd}$}}, and \textcolor{myblue}{\textbf{3$^{rd}$}} best results are highlighted in corresponding colors. $\uparrow$ / $\downarrow$ indicate higher / lower is better. \textbf{BG}, \textbf{FG}, \textbf{TP}, and \textbf{CD} denote Background-Guided, Foreground-Guided, Text-Prompted, and Context-Decoupled, respectively. \textbf{*} denotes that CamoAny results are reproduced by RealCamo~\cite{chen2025realcamo}.}
\label{tab:result}
\renewcommand{\arraystretch}{0.96}
\setlength{\tabcolsep}{3pt}
\resizebox{0.96\textwidth}{!}{
\begin{tabular}{l|c|cc|cc|cc|cc}
\hline
\multirow{2}{*}{\textbf{Method}} &
\multirow{2}{*}{\textbf{Type}} &
\multicolumn{2}{c|}{\textbf{Camouflaged Objects}} &
\multicolumn{2}{c|}{\textbf{Salient Objects}} &
\multicolumn{2}{c|}{\textbf{General Objects}} &
\multicolumn{2}{c}{\textbf{Overall}} \\
\cline{3-10}
& &
FID$\downarrow$ & KID$\downarrow$ &
FID$\downarrow$ & KID$\downarrow$ & 
FID$\downarrow$ & KID$\downarrow$ & 
FID$\downarrow$ & KID$\downarrow$\\
\hline

\textbf{AdaIN} $_{\textcolor{purple}{\textbf{ICCV 2017}}}$~\cite{huang2017arbitrary} & \textbf{BG} &
125.16 & 0.0721 & 133.20 & 0.0702 & 136.93 & 0.0714 & 126.94 & 0.0703\\

\textbf{DCI} $_{\textcolor{purple}{\textbf{AAAI 2020}}}~\cite{zhang2020deep}$ & \textbf{BG} &
130.21 & 0.0689  & 134.92 & 0.0665 & 137.99 & 0.0690 & 130.52 & 0.0673 \\

\textbf{LCGNet}$_{\textcolor{purple}{\textbf{TMM 2023}}}$~\cite{li2022location} & \textbf{BG} &
129.80 & 0.0504 & 136.24 & 0.0597 & 132.64 & 0.0548 & 129.88 & 0.0550\\
\hline

\textbf{TFill} $_{\textcolor{purple}{\textbf{CVPR 2022}}}~\cite{zheng2022bridging}$ & \textbf{FG} &
63.74 & 0.0336 &
96.91 & 0.0453 &
122.44 & 0.0747 &
80.39 & 0.0438 \\

\textbf{RePaint-L} $_{\textcolor{purple}{\textbf{CVPR 2022}}}$~\cite{lugmayr2022repaint} & \textbf{FG} &
76.80 & 0.0459 &
114.96 & 0.0497 & 
136.18 & 0.0686 &
96.14 & 0.0498 \\

\textbf{LAKE-RED} $_{\textcolor{purple}{\textbf{CVPR 2024}}}~\cite{zhao2024lake}$ & \textbf{FG} &
39.55 & 0.0212 &
 88.70 &0.0428 & 102.67 & 0.0625 & 
64.27 &  0.0355 \\

\textbf{FACIG} $_{\textcolor{purple}{\textbf{ICME 2025}}}$~\cite{chen2025foreground} & \textbf{FG} &
\textcolor{mygreen}{\textbf{27.61}} & \textcolor{myblue}{\textbf{0.0099}} &
82.23 & 0.0326 &
\textcolor{mygreen}{\textbf{96.94}} & 0.0504 &
\textcolor{mygreen}{\textbf{52.87}} & 0.0229 \\

\textbf{CamoAny}$^{*}_{\textcolor{purple}{\textbf{CVPR 2025}}}$~\cite{das2025camouflage} & \textbf{FG} &
34.41 & 0.0129 &
\textcolor{mygreen}{\textbf{79.38}} & 0.0314 &
\textcolor{myblue}{\textbf{99.06}} & 0.0539 &
55.40 & 0.0248 \\
\hline
\textbf{LDM-T2I} $_{\textcolor{purple}{\textbf{CVPR 2022}}}$~\cite{rombach2022high} & \textbf{TP} &
58.65 & 0.0380 &
107.38 & 0.0524 &
129.04 & 0.0748 & 
84.48 & 0.0488 \\
\textbf{ControlNet} $_{\textcolor{purple}{\textbf{ICCV 2023}}}$~\cite{zhang2023adding} & \textbf{TP} &
 39.67 &0.0121& 81.72& \textcolor{myblue}{\textbf{0.0303}}& 102.94& \textcolor{myblue}{\textbf{0.0422}}& 59.52& \textcolor{myblue}{\textbf{0.0227}}\\
\textbf{CT-CIG} $_{\textcolor{purple}{\textbf{AAAI 2026}}}$~\cite{qian2026text} & \textbf{TP} &
\textcolor{myblue}{\textbf{30.59}} & \textcolor{mygreen}{\textbf{0.0085}} &
\textcolor{myblue}{\textbf{81.60}} & \textcolor{myred}{\textbf{0.0230}}&
104.46 & \textcolor{myred}{\textbf{0.0241}} &
\textcolor{myblue}{\textbf{52.88}} & \textcolor{mygreen}{\textbf{0.0169}} \\
\hline

\rowcolor{cyan!10}
\textbf{CamoDreamer} (Ours) & \textbf{CD} &
\textcolor{myred}{\textbf{20.44}} & \textcolor{myred}{\textbf{0.0048}} &
\textcolor{myred}{\textbf{59.69}} & \textcolor{mygreen}{\textbf{0.0254}} &
\textcolor{myred}{\textbf{79.76}} & \textcolor{mygreen}{\textbf{0.0420}} &
\textcolor{myred}{\textbf{37.33}} & \textcolor{myred}{\textbf{0.0159}} \\

\hline

\end{tabular}
}
\end{center}
\end{table*}
\subsection{Target Prior Re-weight Optimization}
Given the extreme size imbalance between camouflage objects and their surroundings, the objective in Eq.~\ref{eq2}, which uniformly minimizes the discrepancy between predicted and injected noise, may hinder foreground generation when the target occupies a small image area. To mitigate this imbalance, we introduce a re-weighting scheme based on the target size prior to assign higher importance to foreground regions, where a re-weighting factor $\omega$ is defined as an exponential function of the ratio between the target area and the image area to modulate the mean-squared error loss contribution:
\begin{equation}
\omega = 1 + \lambda\exp\left(-\frac{\sum_{x,y} M'_{x,y}}{H'* W'}\right),
\end{equation}
\begin{equation}
\label{loss}
\begin{split}
\min_{\theta^{\prime}} \mathcal{L}_{\text{CamoDreamer}}
= \mathbb{E}_{z_{0},\epsilon,t,\mathcal{Y}} \Big[
\omega M' \odot \|\epsilon - \mathcal{G}_{\theta,\theta^{\prime}}(z_t,t,\mathcal{Y},F)\|_2^2 \\
+ (1 - M') \odot \|\epsilon - \mathcal{G}_{\theta,\theta^{\prime}}(z_t,t,\mathcal{Y},F)\|_2^2
\Big],
\end{split}
\end{equation}
where $M' \in \{0,1\}^{H' \times W'}$ denotes the object mask aligned with the noisy latent $z_t$, and $(x,y)$ indexes spatial locations. As the object region decreases, $\omega$ increases to encourage the model to lay greater emphasis on foreground regions, while $\lambda$ controls the relative weighting between foreground and background.

\section{Experiments}
\subsection{Experimental Settings}
\textbf{Datasets and Evaluation Metrics.} Following~\cite{das2025camouflage, qian2026text}, we conduct experiments on the LAKE-RED~\cite{zhao2024lake} dataset, comprising 4,040 training images and 19,419 evaluation images. The training set includes 3,040 images from COD10K~\cite{fan2020camouflaged} and 1,000 images from CAMO~\cite{le2019anabranch}. The evaluation set consists of three subsets, namely Camouflage Objects~\cite{lv2021simultaneously}, open-domain Salient Objects~\cite{wang2017learning}, and General Objects~\cite{lin2014microsoft}, each containing 6,473 images. 
For evaluation, we adopt FID~\cite{binkowski2018demystifying} and KID~\cite{heusel2017gans} to assess the fidelity of generated camouflage images. In addition, we report five widely used COD metrics~\cite{pang2022zoom}, including S-measure ($S_{m}$)~\cite{fan2017structure}, weighted F-measure ($F_{\beta}^{w}$)~\cite{margolin2014evaluate}, MAE, F-measure ($F_{\beta}$)~\cite{achanta2009frequency}, and E-measure ($E_{m}$)~\cite{fan2018enhanced}.

\textbf{Implementation Details.} We initialize CamoDreamer from the Stable Diffusion (v1.5) checkpoint, a text-to-image model pre-trained on LAION-5B~\cite{schuhmann2022laion}. All images are resized to $512\times512$ during training. The knowledge base is constructed from the 4,040 training images of LAKE-RED, from which the top-3 candidates are retrieved as background anchors. 
We fine-tune the model using the AdamW optimizer with a fixed learning rate of $1\times10^{-4}$ for 300 epochs, using a total batch size of 320. The relative weighting $\lambda$ is set to 3. 
For inference, we adopt the EulerDiscreteScheduler with 50 sampling steps and a classifier-free guidance scale of 7.5.
\subsection{Comparison with the State-of-the-arts}
Extensive quantitative and qualitative experiments are conducted on the LAKE-RED~\cite{zhao2024lake} dataset to compare CamoDreamer with 11 SOTA methods, including three background-guided methods (AdaIN~\cite{huang2017arbitrary}, DCI~\cite{zhang2020deep}, and LCGNet~\cite{li2022location}), five foreground-guided methods (TFill~\cite{zheng2022bridging}, Repaint-L~\cite{lugmayr2022repaint}, LAKE-RED~\cite{zhao2024lake}, FACIG~\cite{chen2025foreground}, and CamoAny~\cite{das2025camouflage}), and three text-prompted methods (LDM-T2I~\cite{rombach2022high}, CT-CIG~\cite{qian2026text} and ControlNet~\cite{zhang2023adding}). In line with ~\cite{zhao2024lake, das2025camouflage}, CamoDreamer takes a foreground object image, its binary mask, a concise text prompt and the knowledge base as inputs. We reproduce ControlNet under text-guided settings following~\cite{qian2026text}.

\noindent\textbf{$\bullet$ Qualitative Results.}
Fig.~\ref{results} presents qualitative comparisons of synthetic images conditioned on camouflaged and open-domain objects. Background-guided LCGNet tends to overfit objects to surrounding environments, causing severe foreground distortion and even invisibility (\ie, the transparent chameleon in the 2nd row). Foreground-guided methods, such as LAKE-RED and FACIG, often introduce artifacts and semantically implausible backgrounds (\ie, the black background of the bread in the last row and unnatural green surroundings of the centipede in the 3rd row), leading to unrealistic camouflage contexts.  Although text-prompted CT-CIG generates visually concealed objects with stronger semantic guidance, it suffers from mask misalignment and undesired appearance changes, limiting its applicability to downstream tasks. In contrast, our CamoDreamer generates realistic and coherent camouflage images with high contextual compatibility, preserving target characteristics while maintaining consistent backgrounds.
\begin{table}
\caption{\textbf{Quantitative comparison of COD performance using a pretrained ZoomNet~\cite{pang2022zoom} on synthetic camouflage images generated by different methods.}}
\centering
\resizebox{0.49\textwidth}{!}{
\scriptsize
\begin{tabular}{cc|ccccc}
\toprule[1.0pt]% 
 Dataset Type & Methods &$S_{m} \uparrow$&$E_{m} \uparrow$&$F_\beta \uparrow$&$F_\beta^w \uparrow$&$MAE \downarrow$ \\
\hline
\multirow{4}{*}{Camou.Objects}&LAKE-RED~\cite{zhao2024lake}&0.775&0.836&0.695&0.654&0.055\\
&FACIG~\cite{chen2025foreground}&\textcolor{myblue}{\textbf{0.807}}&\textcolor{myblue}{\textbf{0.863}}&\textcolor{myblue}{\textbf{0.741}}&0.705&0.047\\
&CT-CIG~\cite{qian2026text}&0.676&0.764&0.564&0.508&0.097\\
\rowcolor{cyan!10}
&CamoDreamer (ours)&\textcolor{mygreen}{\textbf{0.850}}&\textcolor{mygreen}{\textbf{0.897}}&\textcolor{mygreen}{\textbf{0.789}}&\textcolor{mygreen}{\textbf{0.765}}&\textcolor{myred}{\textbf{0.034}}\\
\hline
\multirow{4}{*}{Salient Objects}&LAKE-RED~\cite{zhao2024lake}&0.663&0.692&0.531&0.487&0.113\\
&FACIG~\cite{chen2025foreground}&0.696&0.711&0.574&0.537&0.103\\
&CT-CIG~\cite{qian2026text}&0.570&0.608&0.402&0.350&0.154\\
\rowcolor{cyan!10}
&CamoDreamer (ours)&\textcolor{myred}{\textbf{0.877}}&\textcolor{myred}{\textbf{0.905}}&\textcolor{myred}{\textbf{0.842}}&\textcolor{myred}{\textbf{0.822}}&\textcolor{mygreen}{\textbf{0.036}}\\
\hline
\multirow{4}{*}{General Objects}&LAKE-RED~\cite{zhao2024lake}&0.606&0.627&0.430&0.390&0.133\\
&FACIG~\cite{chen2025foreground}&0.626&0.631&0.459&0.422&0.127\\
&CT-CIG~\cite{qian2026text}&0.499&0.543&0.278&0.233&0.179\\
\rowcolor{cyan!10}
&CamoDreamer (ours)&0.805&0.824&0.737&\textcolor{myblue}{\textbf{0.709}}&\textcolor{myblue}{\textbf{0.062}}\\
\bottomrule[1.0pt]
\end{tabular}
}
\label{coddetection}
\end{table}

\noindent\textbf{$\bullet$ Quantitative Results.} 
Tab.~\ref{tab:result} reports quantitative evaluations on three test splits of the LAKE-RED dataset. 
Background-guided methods perform the worst due to severe object distortion and large distribution gaps. Foreground-guided methods preserve object structures but suffer from background artifacts and appearance discrepancies, resulting in suboptimal FID/KID scores. Text-prompted methods further improve generation by introducing rich text semantics. Nevertheless, their implicit alignment among text, masks, and image content still causes spatial inconsistencies. 
Overall, CamoDreamer consistently achieves the best performance across all object types, obtaining the lowest FID (37.33) and KID (0.0159), with a 15.54 improvement over the previous best FID. It also reduces the performance gap between camouflaged and salient/general objects, highlighting strong camouflage transfer capability.
\begin{table}
\caption{Comparison of Trainable and Total Parameters between CamoDreamer and Diffusion-based CIG Methods.}
\centering
\resizebox{0.46\textwidth}{!}{
\scriptsize
\begin{tabular}{c|ccc}
\toprule[1.0pt]% 
 Method & Trainable Params (M)$\downarrow$ &Overall params (M) $\downarrow$& FID $\downarrow$ \\
\hline
LAKE-RED~\cite{zhao2024lake} & \textcolor{myblue}{\textbf{387.25}}& \textcolor{myred}{\textbf{440.47}}&64.27\\
ControlNet~\cite{zhang2023adding}&1,230.14& \textcolor{myblue}{\textbf{1,667.83}}&\textcolor{myblue}{\textbf{59.52}}\\
CT-CIG~\cite{qian2026text}&\textcolor{myred}{\textbf{106.22}}& 3472.57& \textcolor{mygreen}{\textbf{52.88}}\\
\rowcolor{cyan!10}
CamoDreamer (ours)&\textcolor{mygreen}{\textbf{172.22}}&\textcolor{mygreen}{\textbf{1238.45}}&\textcolor{myred}{\textbf{37.33}}\\
\bottomrule[1.0pt]
\end{tabular}
}
\label{computation}
\end{table}
\begin{figure}[!t]
\centering
\includegraphics[width=0.48\textwidth]{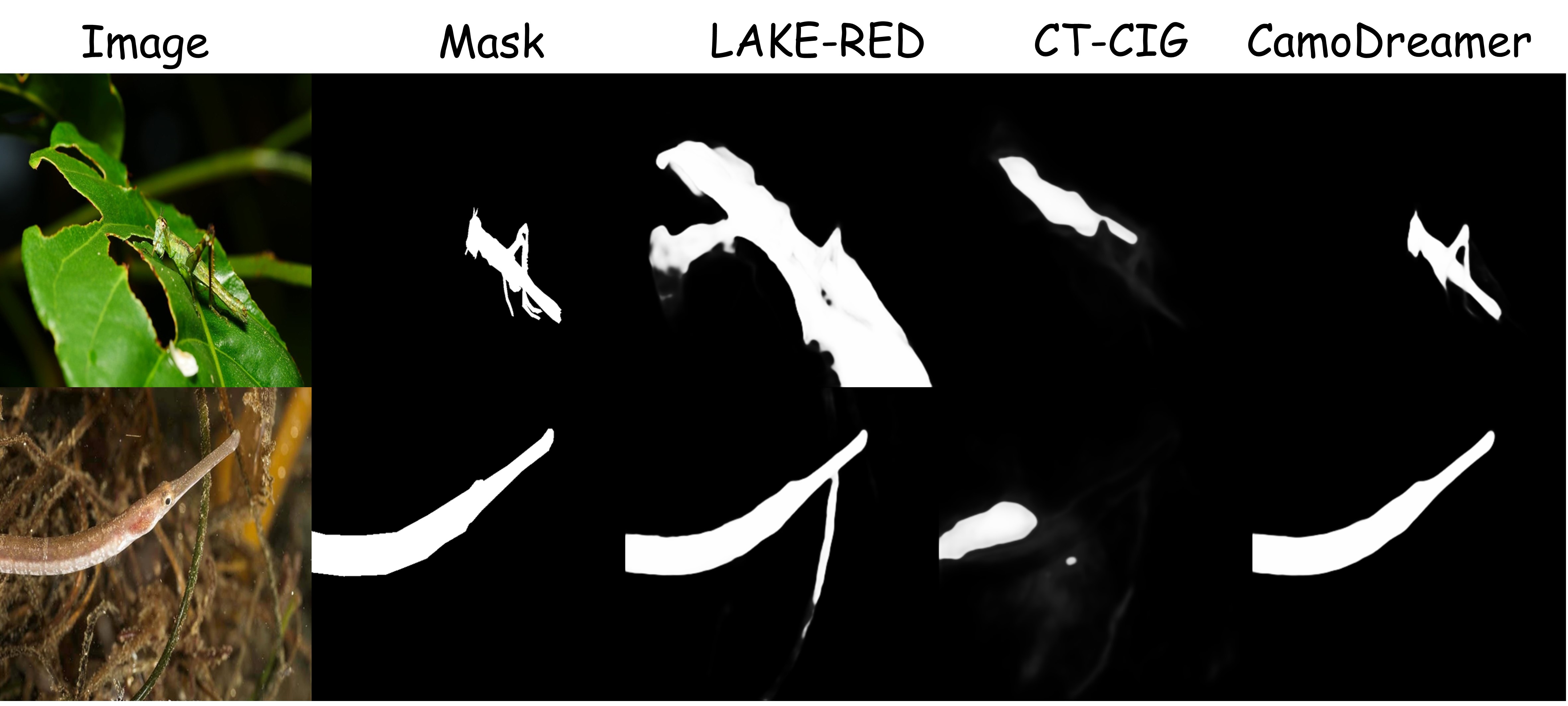}
\caption{ZoomNet~\cite{pang2022zoom} predictions on the COD10K dataset when trained with different synthesized camouflage images.}
\label{codresults}
\end{figure}

To further evaluate structural alignment, we employ a pretrained COD model (\ie, ZoomNet~\cite{pang2022zoom}) to predict object masks and compare them with ground-truth annotations. Higher detection scores (\ie, $S_m$, $E_m$, $F_\beta$, and $F_\beta^w$) indicate better structural preservation and shape alignment, while lower MAE reflects accurate object localization. As shown in Tab.~\ref{coddetection}, CamoDreamer consistently achieves superior performance across all settings, demonstrating its ability to maintain object structures without distortion. Moreover, Tab.~\ref{computation} compares the trainable and total parameters of CamoDreamer with other diffusion-based CIG methods. Parameter differences mainly arise from the choices of VAE, text encoder, and U-Net backbone architectures. Despite achieving superior fidelity, CamoDreamer is deployed only on the U-Net middle and decoder layers, maintaining a relatively low number of trainable parameters (172.22M), significantly fewer than ControlNet and competitive with prior methods, achieving a favorable balance between training efficiency and generation quality.

\noindent\textbf{$\bullet$ Downstream Utility.}
To further assess the impact of generated camouflage images on downstream perception, we train the COD model ZoomNet~\cite{pang2022zoom} using synthesized LAKE-RED training sets (\ie, 4,040 images) and evaluate it on three benchmarks (CAMO, COD10K, and NC4K). Quantitative and qualitative results are reported in Tab.~\ref{tab:COD} and Fig.~\ref{codresults}. We observe that training on data synthesized by CamoDreamer yields the best performance across all datasets and metrics. In particular, it surpasses CT-CIG by a large margin (\ie, +0.180 $S_m$ on NC4K), and further outperforms LAKE-RED with consistent gains in $S_m$ (+0.053 / +0.081 / +0.049 on CAMO, COD10K, and NC4K), along with improvements across other metrics. These results indicate that our synthesized data effectively narrows the domain gap to real-world distributions, delivering superior downstream detection performance.

\begin{table*}[htbp]
\begin{center}
\caption{\textbf{Performance of ZoomNet~\cite{pang2022zoom} trained on different synthesized camouflage datasets on three COD benchmarks.}
 Syn. (Synthesized) denotes synthetic camouflage images generated by different CIG methods. }
\label{tab:COD}
\renewcommand{\arraystretch}{1.0}
\setlength{\tabcolsep}{3pt}
\resizebox{1.0\textwidth}{!}{
\begin{tabular}{l|ccccc|ccccc|ccccc}
\hline
\multirow{2}{*}{\textbf{Training Data}} &
\multicolumn{5}{c|}{\textbf{CAMO (250)}} &
\multicolumn{5}{c|}{\textbf{COD10K (2,026)}} &
\multicolumn{5}{c}{\textbf{NC4K (4,121)}} \\
\cline{2-16}
&
$S_{m} \uparrow$&$E_{m} \uparrow$&$F_\beta \uparrow$&$F_\beta^w \uparrow$&$MAE \downarrow$&
$S_{m} \uparrow$&$E_{m} \uparrow$&$F_\beta \uparrow$&$F_\beta^w \uparrow$&$MAE \downarrow$&
$S_{m} \uparrow$&$E_{m} \uparrow$&$F_\beta \uparrow$&$F_\beta^w \uparrow$&$MAE \downarrow$\\
\hline

Syn.LAKE-RED &
\textcolor{mygreen}{\textbf{0.649}} & \textcolor{mygreen}{\textbf{0.693}} & \textcolor{mygreen}{\textbf{0.582}} & \textcolor{mygreen}{\textbf{0.530}} & 0.179 &
\textcolor{mygreen}{\textbf{0.676}} & \textcolor{mygreen}{\textbf{0.710}} & \textcolor{mygreen}{\textbf{0.513}} & \textcolor{mygreen}{\textbf{0.469}} & 0.121 &
\textcolor{mygreen}{\textbf{0.746}} & \textcolor{mygreen}{\textbf{0.787}} & \textcolor{mygreen}{\textbf{0.664}} & \textcolor{mygreen}{\textbf{0.620}} & \textcolor{mygreen}{\textbf{0.105}} \\

Syn.CT-CIG &
0.574 & 0.592 & 0.428 & 0.357 & \textcolor{mygreen}{\textbf{0.143}} &
0.604 & 0.656 & 0.392 & 0.328 & \textcolor{mygreen}{\textbf{0.078}} &
0.615 & 0.651 & 0.482 & 0.407 & 0.116 \\

\rowcolor{cyan!10}
Syn.CamoDreamer  &
\textcolor{myred}{\textbf{0.702}} & \textcolor{myred}{\textbf{0.725}} & \textcolor{myred}{\textbf{0.611}} & \textcolor{myred}{\textbf{0.564}} & \textcolor{myred}{\textbf{0.102}} &
\textcolor{myred}{\textbf{0.757}}&\textcolor{myred}{\textbf{0.780}}&\textcolor{myred}{\textbf{0.644}}&\textcolor{myred}{\textbf{0.596}}&\textcolor{myred}{\textbf{0.045}} &
\textcolor{myred}{\textbf{0.795}}&\textcolor{myred}{\textbf{0.826}}&\textcolor{myred}{\textbf{0.740}}&\textcolor{myred}{\textbf{0.697}}&\textcolor{myred}{\textbf{0.060}} \\
\hline
\end{tabular}}
\end{center}
\end{table*}
\begin{table*}[t]
\centering
\caption{Ablation studies on two key components of the proposed module (\ie, contrast embedding and the assimilation branch), as well as the target prior re-weight optimisation.}
\begin{subtable}[t]{0.32\textwidth}
\centering
\renewcommand{\arraystretch}{0.8}
\caption{Ablation of contrast embedding in CCB.}
\resizebox{0.9\textwidth}{!}{
\begin{tabular}{c|cc}
\toprule
\multirow{2}*{\makecell{Contrast\\ Embedding}} &\multicolumn{2}{c}{Overall} \\
\cmidrule{2-3}
&FID$\downarrow$&KID$\downarrow$\\

\midrule
$\times$ &41.74&0.0188\\
$\checkmark$ &37.33 &0.0159\\
\bottomrule
\end{tabular}
}
\end{subtable}
\hfill
\begin{subtable}[t]{0.32\textwidth}
\centering
\renewcommand{\arraystretch}{0.8}
\caption{Ablation of assimilation branch in CDAS.}
\resizebox{0.9\textwidth}{!}{
\begin{tabular}{c|cc}
\toprule
\multirow{2}*{\makecell{Assimilation\\ Branch}} &\multicolumn{2}{c}{Overall} \\
\cmidrule{2-3}
&FID$\downarrow$&KID$\downarrow$\\

\midrule
$\times$& 42.29&0.0181\\
$\checkmark$&37.33 &0.0159 \\
\bottomrule
\end{tabular}
}
\end{subtable}
\hfill
\begin{subtable}[t]{0.32\textwidth}
\centering
\renewcommand{\arraystretch}{0.8}
\caption{Ablation of target prior re-weight.}
\resizebox{0.9\textwidth}{!}{
\begin{tabular}{c|cc}
\toprule
\multirow{2}*{\makecell{Target Prior\\ Re-weight}} &\multicolumn{2}{c}{Overall} \\
\cmidrule{2-3}
&FID$\downarrow$&KID$\downarrow$\\

\midrule
$\times$& 46.84&0.0181\\
$\checkmark$&37.33 &0.0159 \\
\bottomrule
\end{tabular}
}
\end{subtable}
\label{tab:three_subtables}
\end{table*}

\begin{table}[t]
\centering
\renewcommand{\arraystretch}{0.8}
\caption{\textbf{Quantitative ablation study of each component in CamoDreamer.} We progressively remove individual modules from the full model to evaluate their contributions to CIG. The results demonstrate the effectiveness of each component.}
\label{tab:ablation}
\resizebox{0.5\textwidth}{!}{
\begin{tabular}{ccccccc}
\toprule
\multicolumn{3}{c}{\textbf{Module}} &\multicolumn{2}{c}{Computation Cost} & \multicolumn{2}{c}{Overall} \\
\cmidrule(lr){1-3} \cmidrule(lr){4-5} \cmidrule(lr){6-7}
CCB & CDAS & FACB& Trainable Params (M) & Infer. Time (s/img)& FID$\downarrow$ & KID$\downarrow$ \\
\midrule
$\times$ & $\times$ & $\times$ & 56.49 &4.0267&60.85&0.0305\\
$\times$ & $\checkmark$ & $\times$ & 97.44 & 5.2164& 55.31 & 0.0251 \\
$\times$ & $\checkmark$ & $\checkmark$ &108.31& 6.0608 &51.60 & 0.0236 \\
$\checkmark$ & $\checkmark$ & $\times$ &  161.35 & 5.3514 & 46.52 & 0.0180 \\
$\checkmark$ & $\checkmark$ & $\checkmark$ &  172.22&6.1872 & 37.33& 0.0159 \\
\bottomrule[1.0pt]
\end{tabular}
} 
\end{table}
\begin{figure}[!t]
\centering
\includegraphics[width=0.46\textwidth]{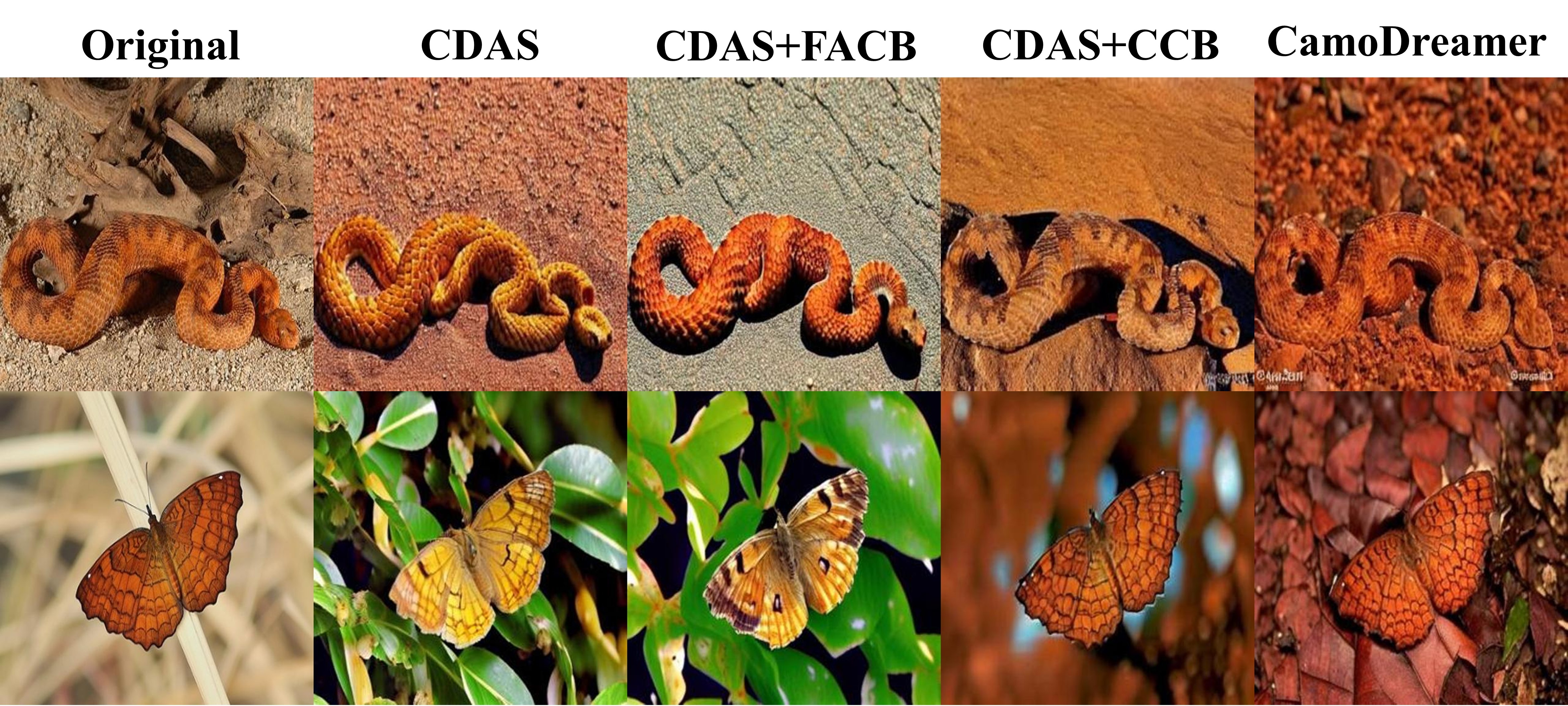}
\caption{Visualisations of ablation studies on the proposed components in our CamoDreamer.}
\label{ablation}
\end{figure}
\subsection{Ablation Study}
We conduct comprehensive ablations to evaluate the contribution of each component in CamoDreamer. As shown in Fig.~\ref{ablation}, CDAS alone achieves reasonable generation, but produces coarse contextual camouflage and object integrity due to the absence of CCB, which limits effective decoupling and assimilation. Quantitatively, Tab.~\ref{tab:three_subtables}(b) shows that introducing the assimilation branch reduces FID from 42.29 to 37.33, indicating improved background rendering.  

Building upon CDAS, CCB further enhances camouflage realism by capturing cross-context inconsistencies. In Tab.~\ref{tab:three_subtables}(a), contrast embedding reduces FID from 41.74 to 37.33 by improving texture consistency and suppressing artifacts. This is validated in Tab.~\ref{tab:ablation}, where removing CCB degrades FID from 37.33 to 51.60, underscoring its importance in modeling subtle foreground–background discrepancies. Fig.~\ref{ablation}  illustrates that CCB improves contextual compatibility. 
Moreover, introducing FACB completes the frequency-aware modulation, yielding the best performance (FID: 37.33 \textit{vs.} 46.52 without FACB) by enhancing fine-grained textures and structural fidelity. Finally, the target prior re-weight further stabilizes generation by emphasizing small foreground objects, reducing FID from 46.84 to 37.33 (Tab.~\ref{tab:three_subtables}(c)). Overall, both qualitative results in Fig.~\ref{ablation}, where the snake and butterfly exhibit progressively stronger camouflage, and consistent quantitative gains demonstrate that each component contributes complementary benefits, with their integration achieving the best camouflage results. More hyperparameter ablations are provided in the appendix.

\section{Conclusion and Future Work}
In this paper, we propose the first context-decoupled camouflage generation paradigm, termed CamoDreamer. To address cross-context representation leakage in existing CIG methods, we introduce a ``decouple-then-blend'' paradigm that explicitly separates camouflage features into collaborative object–background control streams for coherent contextual rendering. 
CamoDreamer consists of three key components: (1) a Contrast-aware Contextual Bridge that models target–background associations to construct dual conditional guidance; (2) Context-Decoupled Assimilation Streams that independently render foreground and background; and (3) a Frequency-Adaptive Context Blend module that integrates decoupled features into a unified camouflage representation. 
Extensive experiments demonstrate the superior fidelity of our CamoDreamer over existing CIG methods. 
In the future, we plan to extend the context-decoupled paradigm to controllable camouflage video generation.

\begin{acks}
This work is partially supported by grants from the National Natural Science Foundation of China (No.62132002), Guizhou Provincial Major Scientific and Technological Program (Qiankehe Zhongda [2025] No. 032), Beijing Nova Program (No.20250484786), Shandong Provincial Natural Science Foundation (No. ZR2026LZJ009), and the Fundamental Research Funds for the Central Universities.
\end{acks}

\bibliographystyle{ACM-Reference-Format}
\bibliography{main}

\clearpage
%\newpage
\appendix
%\onecolumn
\begin{center}
\Huge \textbf{Supplementary Materials}
\end{center}
This supplementary material provides additional experimental details and analyses omitted from the main paper due to space limitations. To further evaluate the proposed CamoDreamer for camouflage image generation (CIG), it includes: (1) detailed descriptions of the datasets and implementation settings; and (2) extensive qualitative and quantitative ablation studies with corresponding analyses.
\section{Dataset and Implementation Details}
Following existing CIG methods~\cite{zhao2024lake,das2025camouflage,qian2026text}, we conduct experiments on the LAKE-RED~\cite{zhao2024lake} dataset, which contains 4,040 training images and 19,419 evaluation images. As described in the main paper, the evaluation set consists of three subsets: 6,473 camouflage images, 6,473 open-domain salient images, and 6,473 general images from MS-COCO~\cite{lin2014microsoft}. Following the standard evaluation protocol, FID and KID are computed by comparing generated images with 5,066 real camouflage images from the COD10K (CAM) subset of LAKE-RED. CamoDreamer is deployed on the middle (\ie, $8\times8$) and decoder (\ie, $16\times16$, $32\times32$, and $64\times64$) layers of SDv1.5~\cite{rombach2022high} and optimized with FP16, enabling training on a single RTX 3090 GPU (24GB). Following prior works~\cite{das2025camouflage,qian2026text}, we employ Qwen2-VL~\cite{wang2024qwen2} to generate concise text descriptions for each image, \ie, ``Camouflaged fish in an underwater rocky-sandy habitat.''
\section{More Quantitative and Qualitative Results}
\subsection{Quantitative Results}
\noindent\textbf{$\bullet$ Ablation on the weighting factor $\lambda$.} For the target-prior re-weighting strategy, Tab.~\ref{lambda} shows the effect of varying $\lambda$. A moderate value ($\lambda=3$) achieves the best performance. Smaller values (\ie, $\lambda=2$) underutilize target-aware guidance, leading to weaker camouflage alignment, whereas larger values (\ie, $\lambda=4$) overemphasize target cues and disrupt natural blending with the background.

\noindent\textbf{$\bullet$ Ablation on the frequency components}. For the frequency-adaptive contextual blend (FACB), we further analyze the contributions of high- and low-frequency components. As shown in Tab.~\ref{freq.}, using only low-frequency information significantly improves performance over the baseline (\ie, without frequency modeling), reducing the overall FID from 46.52 to 41.02. This indicates that low-frequency components effectively capture global image structures and coarse semantic content, which are essential for maintaining camouflage consistency. Building on this, incorporating high-frequency components further improves performance, achieving the best results (FID 37.33 and KID 0.0159). This demonstrates that high-frequency details, such as fine-grained object textures and boundary patterns, further enhance local realism and visual fidelity. Collectively, these results demonstrate that FACB effectively integrates complementary high- and low-frequency content.
\begin{figure}[!t]
\centering
\includegraphics[width=0.45\textwidth]{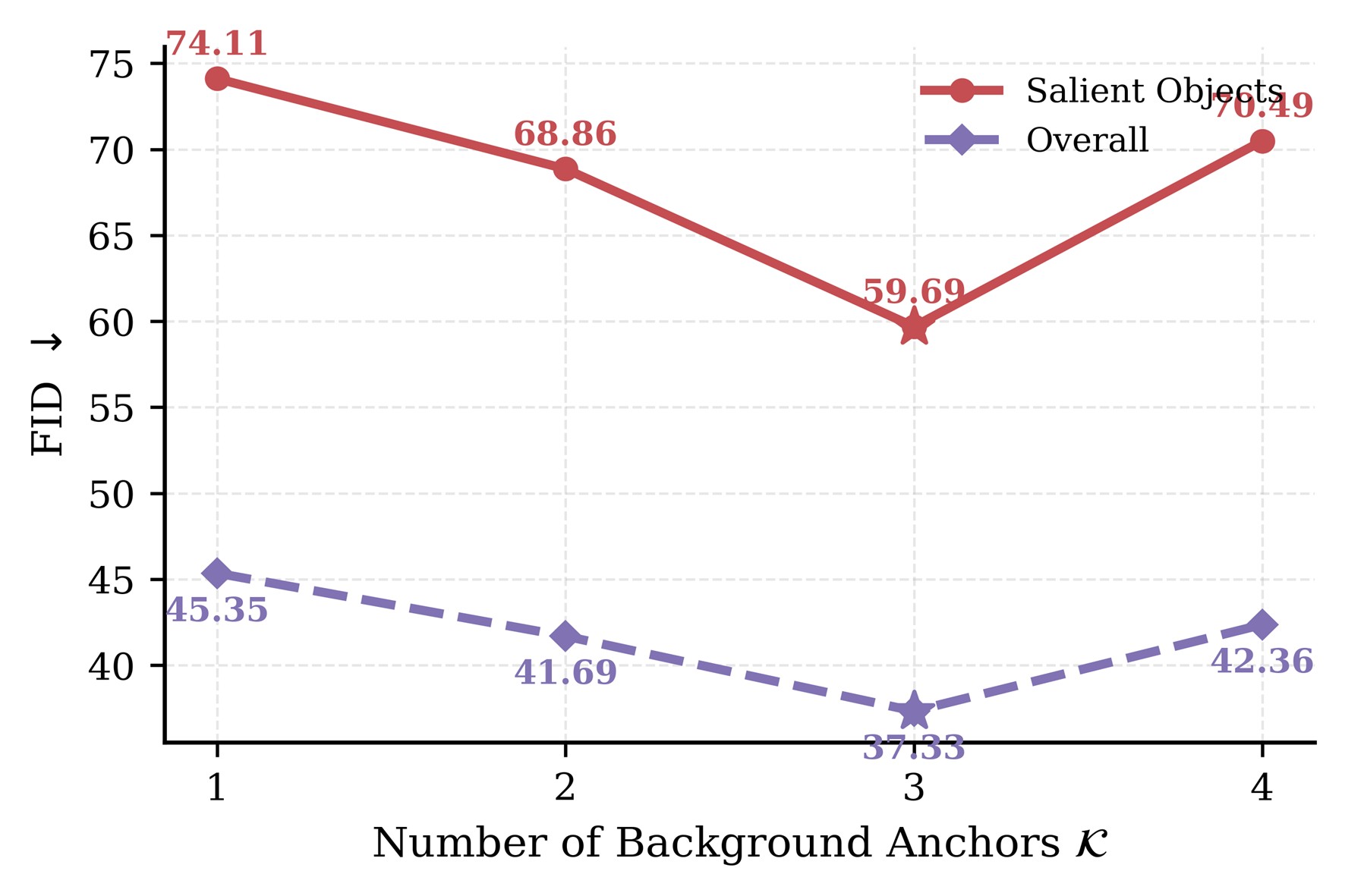}
\caption{Qualitative ablation on the number of background anchors (\ie, Top-$\mathcal{K}$) in CCB.}
\label{topk}
\end{figure}
\begin{table}[t]
\centering
\renewcommand{\arraystretch}{0.6}
\caption{\textbf{Quantitative ablation study of the relative weighting $\lambda$ in the optimization objective.} }
\label{lambda}
\resizebox{0.4\textwidth}{!}{
\begin{tabular}{ccccc}
\toprule
\multirow{2}*{$\lambda$} &\multicolumn{2}{c}{\textbf{Camou. Objects}} & \multicolumn{2}{c}{\textbf{Overall}} \\
 \cmidrule(lr){2-3} \cmidrule(lr){4-5}
& FID$\downarrow$ & KID$\downarrow$ & FID$\downarrow$ & KID$\downarrow$ \\
\midrule
$1$  &  20.50& 0.0050 & 37.67& 0.0162 \\
$2$ &  19.57 & 0.0051 & 45.24 & 0.0182 \\
$3$ & 20.44& 0.0048 & 37.33 & 0.0159 \\
$4$ &20.10 & 0.0056 & 43.46 & 0.0189 \\
\bottomrule
\end{tabular}
}
\end{table}
\begin{table}[t]
\centering
\renewcommand{\arraystretch}{0.6}
\caption{\textbf{Quantitative ablation study of high-/low-frequency components in FACB.} }
\label{freq.}
\resizebox{0.4\textwidth}{!}{
\begin{tabular}{cccccc}
\toprule
\multicolumn{2}{c}{\textbf{Freq.}}&\multicolumn{2}{c}{\textbf{Camou. Objects}} & \multicolumn{2}{c}{\textbf{Overall}} \\
\cmidrule(lr){1-2} \cmidrule(lr){3-4} \cmidrule(lr){5-6}
high&low& FID$\downarrow$ & KID$\downarrow$ & FID$\downarrow$ & KID$\downarrow$ \\
\midrule
$\times$ & $\times$  &  24.32 &0.0067& 46.52 &0.0180 \\
$\times$ & $\checkmark$ &  19.13 & 0.0047 & 41.02 & 0.0167 \\
$\checkmark$ & $\checkmark$ & 20.44& 0.0048 & 37.33 &0.0159 \\
\bottomrule
\end{tabular}
}
\end{table}

Moreover, Tab.~\ref{tab:three_subtables} provides additional ablation studies on key design choices in CamoDreamer, including the number of background anchors, foreground attention (Eq.~\ref{foreattention} in the main paper), feature extractors, and the learnable frequency masks in FACB. 

\noindent\textbf{$\bullet$ Ablation on the background anchors.} As shown in Tab.~\ref{tab:three_subtables}(a), removing background anchors significantly degrades FID and KID, indicating that foreground-only modeling is insufficient for realistic camouflage generation. Without background references, the model cannot effectively capture contextual information, resulting in reduced contextual coherence and camouflage realism.
\begin{figure}[!t]
\centering
\includegraphics[width=0.49\textwidth]{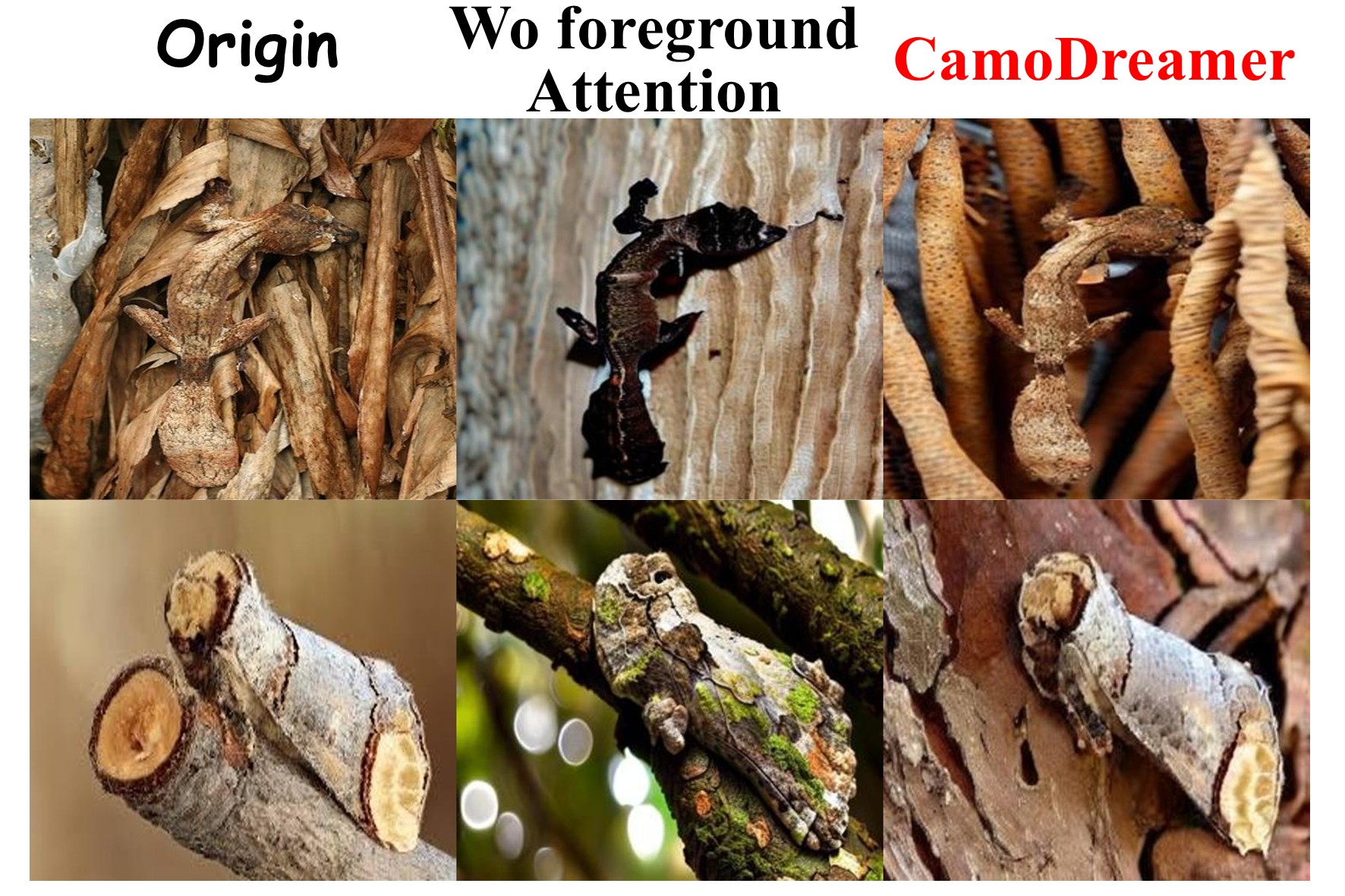}
\caption{Qualitative ablation of foreground attention in the object control stream.}
\label{foattention}
\end{figure}

\noindent\textbf{$\bullet$ Ablation on the foreground attention.} For the foreground attention in the object control stream, removing this module causes a significant performance drop (\ie, FID 51.55 \textit{vs.} 37.33 and KID 0.0236 \textit{vs.} 0.0159 in Tab.~\ref{tab:three_subtables} (b)). This demonstrates that foreground-guided feature interaction is essential for preserving target characteristics and improving camouflage generation quality.

\noindent\textbf{$\bullet$ Ablation on the feature extractors.} For feature extraction, replacing ConvNeXt-Tiny~\cite{liu2022convnet} with CLIP~\cite{radford2021learning} image encoders results in a notable performance drop (\ie, FID 49.24 \textit{vs.} 37.33 in Tab.~\ref{tab:three_subtables} (c)). Although CLIP encoders provide powerful general-purpose visual representations, their pre-trained features are not sufficiently aligned with camouflage-specific patterns. Meanwhile, fully fine-tuning large-scale vision-language models ($\sim304$M) introduces substantial computational overhead. Therefore, the lightweight ConvNeXt-Tiny ($\sim28$M) offers a more efficient and effective solution for capturing domain-specific camouflage features.

\noindent\textbf{$\bullet$ Ablation on the learnable frequency mask.}
In addition, we investigate the effectiveness of the learnable frequency mask in FACB. For the fixed variant, frequency components are separated into high- and low-frequency bands using a predefined cutoff threshold of 0.5 based on the normalized radius. As shown in Tab.~\ref{tab:three_subtables} (d), replacing the learnable mask with the fixed one degrades performance (FID: 37.33 $\rightarrow$ 39.87), demonstrating the necessity of adaptive frequency decomposition. Unlike manually defined thresholds, the learnable mask dynamically adjusts frequency selection according to image content, enabling FACB to better preserve low-frequency structures while enhancing high-frequency object details, thereby achieving more coherent foreground-background camouflage synthesis.
\begin{table*}[t]
\centering
\caption{Ablation studies on key design choices, including background anchors, foreground attention, feature extractors, and the learnable frequency mask in FACB.}
\begin{subtable}[t]{0.24\textwidth}
\centering
\renewcommand{\arraystretch}{0.8}
\caption{Ablation of background anchors.}
\resizebox{0.9\textwidth}{!}{
\begin{tabular}{c|cc}
\toprule
\multirow{2}*{\makecell{Background\\ Anchors}} &\multicolumn{2}{c}{Overall} \\
\cmidrule{2-3}
&FID$\downarrow$&KID$\downarrow$\\

\midrule
$\times$ &50.24&0.0197\\
$\checkmark$ &37.33 &0.0159\\
\bottomrule
\end{tabular}
}
\end{subtable}
\hfill
\begin{subtable}[t]{0.24\textwidth}
\centering
\renewcommand{\arraystretch}{0.8}
\caption{Ablation of foreground attention.}
\resizebox{0.9\textwidth}{!}{
\begin{tabular}{c|cc}
\toprule
\multirow{2}*{\makecell{Foreground\\ Attention}} &\multicolumn{2}{c}{Overall} \\
\cmidrule{2-3}
&FID$\downarrow$&KID$\downarrow$\\

\midrule
$\times$ &51.55 & 0.0236\\
$\checkmark$ &37.33 &0.0159\\
\bottomrule
\end{tabular}
}
\end{subtable}
\hfill
\begin{subtable}[t]{0.24\textwidth}
\centering
\renewcommand{\arraystretch}{0.8}
\caption{Ablation of Feature Extractors in CCB.}
\resizebox{1.0\textwidth}{!}{
\begin{tabular}{c|cc}
\toprule
\multirow{2}*{\makecell{Feature\\ Extractors}} &\multicolumn{2}{c}{Overall} \\
\cmidrule{2-3}
&FID$\downarrow$&KID$\downarrow$\\

\midrule
$\text{CLIP}$& 49.24&0.0227\\
$\text{ConvNeXt-Tiny}$&37.33 &0.0159 \\
\bottomrule
\end{tabular}
}
\end{subtable}
\hfill
\begin{subtable}[t]{0.24\textwidth}
\centering
\renewcommand{\arraystretch}{0.8}
\caption{Ablation of learnable frequency mask.}
\resizebox{1.0\textwidth}{!}{
\begin{tabular}{c|cc}
\toprule
\multirow{2}*{\makecell{learnable\\ frequency mask}} &\multicolumn{2}{c}{Overall} \\
\cmidrule{2-3}
&FID$\downarrow$&KID$\downarrow$\\

\midrule
$\times$& 39.87&0.0175\\
$\checkmark$&37.33 &0.0159 \\
\bottomrule
\end{tabular}
}
\end{subtable}
\label{tab:three_subtables}
\end{table*}
\begin{figure}[!t]
\centering
\includegraphics[width=0.48\textwidth]{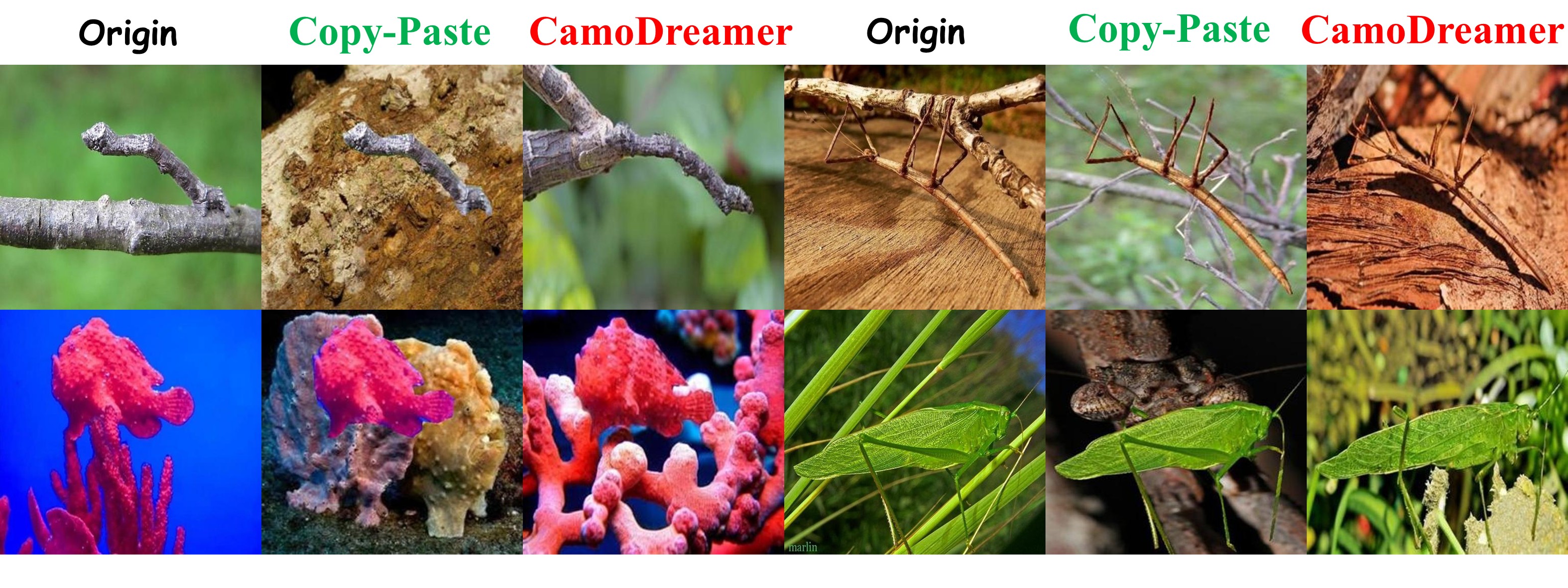}
\caption{Qualitative comparison between CamoDreamer and a non-generative method (\ie, Copy-Paste).}
\label{copypaste}
\end{figure}
\subsection{Qualitative Results}
Fig.~\ref{topk} presents qualitative comparisons with different numbers of background anchors ($\mathcal{K}$) in CCB. Increasing $\mathcal{K}$ from 1 to 3 consistently improves the FID score on both salient-object and overall evaluations, demonstrating that multiple candidate backgrounds provide richer contextual priors and facilitate more effective foreground-background alignment. The best performance is achieved at $\mathcal{K}=3$ (overall FID 37.33), whereas further increasing $\mathcal{K}$ to 4 results in performance degradation. This indicates that excessive background anchors may introduce redundant or semantically inconsistent context, weakening contextual compatibility during camouflage synthesis.

Fig.~\ref{foattention} demonstrates that removing the attention mechanism in the object control stream compromises target fidelity and camouflage compatibility (\ie, the black lizard and green wood cases). Furthermore, Fig.~\ref{copypaste} presents a qualitative comparison between CamoDreamer and the non-generative Copy-Paste method. We observe that Copy-Paste suffers from evident integration artifacts and appearance inconsistencies, as it directly composites objects without context modeling, resulting in unrealistic camouflage patterns. This limitation is also reflected in the quantitative results on COD objects (6,473 images), where Copy-Paste obtains FID/KID scores of \textbf{38.28/0.0133}, compared with \textbf{20.44/0.0048} for CamoDreamer.

\begin{figure*}[!t]
\centering
\includegraphics[width=1.00\textwidth]{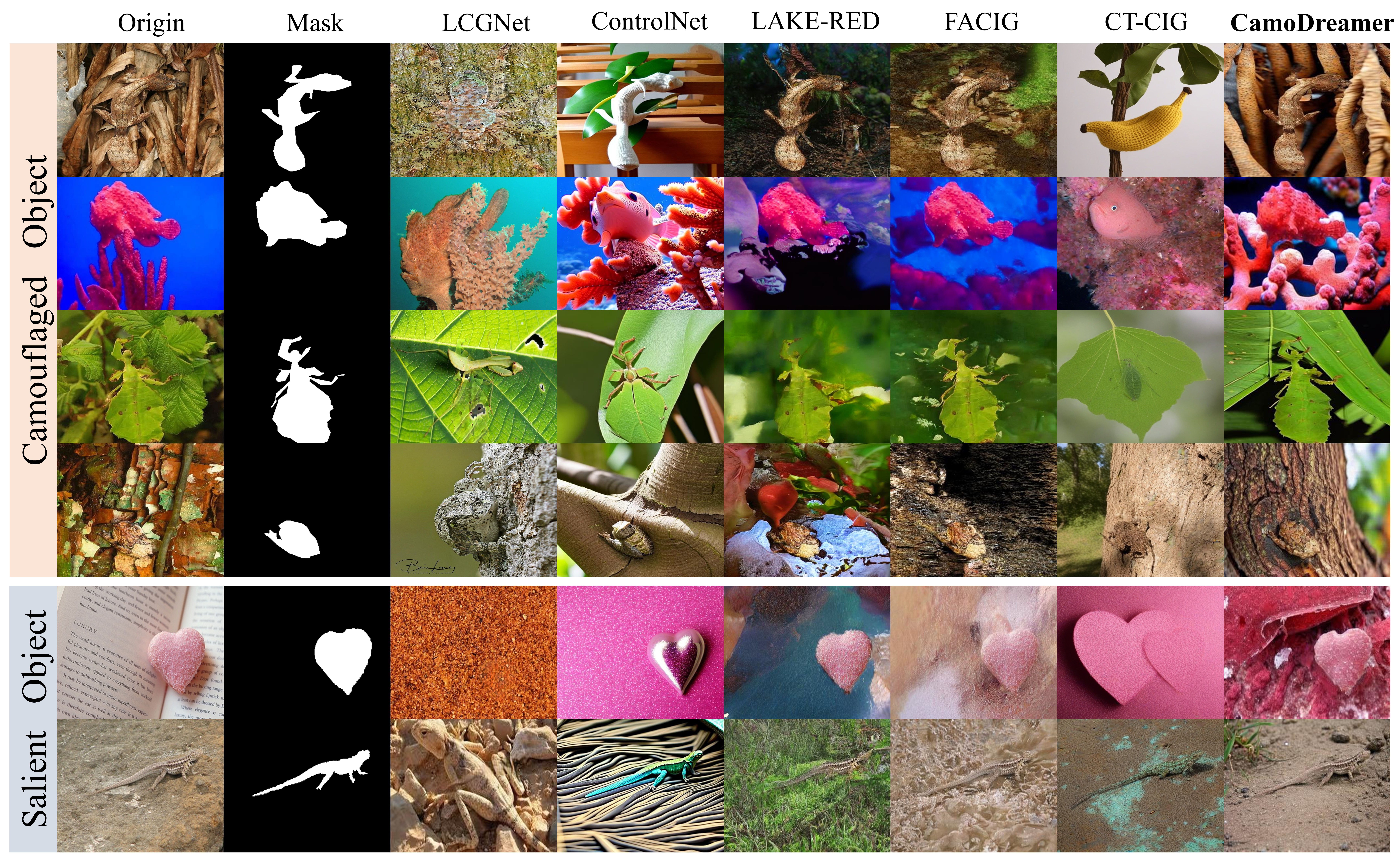}
\caption{Additional qualitative comparisons of camouflage images generated by different CIG methods.}
\label{supple_ablation_compare}
\end{figure*}

Fig.~\ref{supple_ablation_compare} provides additional qualitative comparisons with existing CIG methods. As observed, existing approaches often face challenges in balancing camouflage compatibility and object fidelity. Text-prompted methods, such as ControlNet~\cite{zhang2023adding} and CT-CIG~\cite{qian2026text}, preserve object semantics through textual guidance but frequently suffer from appearance inconsistencies (\ie, the white chameleon and banana in the 1st row). Meanwhile, foreground-guided methods, such as LAKE-RED~\cite{zhao2024lake} and FACIG~\cite{chen2025foreground}, focus on object preservation but may introduce unrealistic background artifacts (\ie, the false background around the green mantis in the 3rd row), leading to degraded contextual coherence. 
Benefiting from the proposed context-decoupled paradigm, CamoDreamer consistently generates camouflage images with improved foreground-background compatibility. Specifically, it preserves target structural integrity while adapting surrounding contexts, enabling coherent integration between objects and diverse environments. This advantage is particularly pronounced in challenging scenarios with complex textures and heterogeneous backgrounds, where maintaining both object fidelity and contextual consistency is critical.

\begin{figure*}[!t]
\centering
\includegraphics[width=0.91\textwidth]{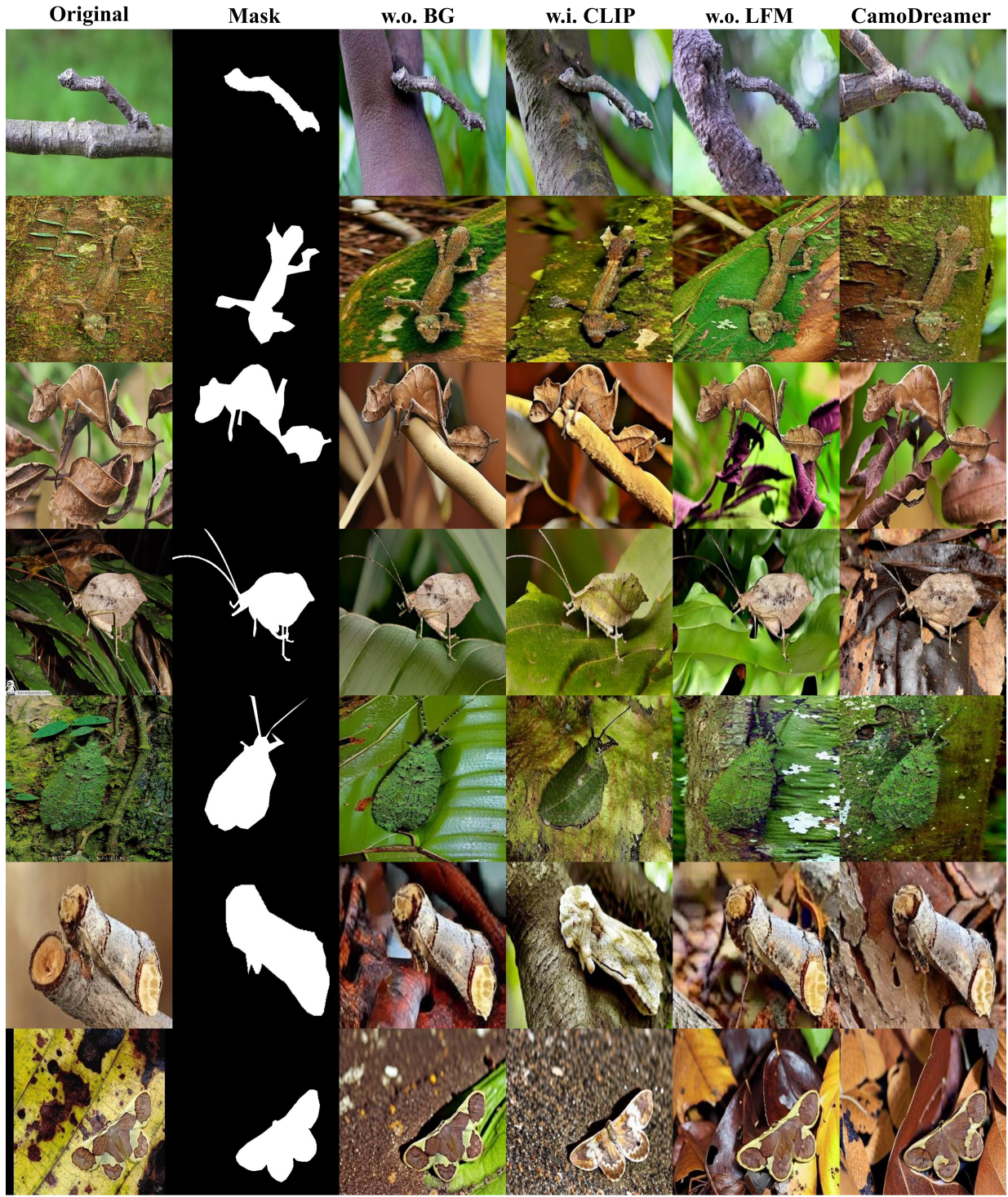}
\caption{Qualitative visualisations of ablation studies on the proposed components in our CamoDreamer, encompassing background anchors (BG), feature extractors (\ie, CLIP), and learnable frequency masks (LFM).}
\label{supple_ablation}
\end{figure*}

Fig.~\ref{supple_ablation} further presents qualitative comparisons for the ablation studies of key components in CamoDreamer. Removing background anchors (\ie, w.o. BG) substantially degrades visual quality, as the model lacks contextual priors from surrounding environments and fails to establish effective foreground-background integration, often producing isolated objects with limited camouflage compatibility. Replacing the ConvNeXt-Tiny feature extractor with CLIP (w.i. CLIP) introduces noticeable texture and color inconsistencies, suggesting that CLIP’s general-purpose representations are insufficiently aligned with domain-specific camouflage characteristics. 

In addition, removing the learnable frequency mask (w.o. LFM) weakens the adaptive fusion of structural and textural information, leading to either over-smoothed structures or inconsistent high-frequency details. In contrast, the full CamoDreamer model achieves more balanced and visually coherent results by effectively integrating complementary frequency cues. These qualitative observations align with the quantitative results and further demonstrate the contribution of each component to realistic and coherent camouflage synthesis.
\subsection{Generative Limitations}
CamoDreamer adopts a context-decoupled generation paradigm that explicitly disentangles foreground objects from background context and regulates their interactions through separate modeling, effectively alleviating cross-context representation leakage. This design prioritizes contextual compatibility, which is essential for realistic camouflage synthesis.

Despite achieving strong generation performance, CamoDreamer still faces challenges in open-domain scenarios. Specifically, for highly salient or structurally complex objects, the model may occasionally introduce local distortions or compromise fine-grained structural details due to the inherent difficulty of balancing object preservation and environmental adaptation. Compared with foreground-guided or text-prompted approaches that primarily emphasize object fidelity or semantic alignment, CamoDreamer's design focuses on achieving harmonious foreground-background integration. Consequently, it provides an effective balance between camouflage compatibility and target preservation, while maintaining room for further improvement in preserving complex object structures under unconstrained conditions.

\end{document}